\documentclass[Afour,sageh,times]{sagej}

\usepackage{moreverb,url}

\usepackage[colorlinks,bookmarksopen,bookmarksnumbered,linkcolor=blue,citecolor=blue,urlcolor=blue]{hyperref}

\usepackage{bm}
\usepackage{lipsum}
\usepackage{siunitx}
\usepackage{amsthm}
\usepackage[nolist,nohyperlinks]{acronym}
\begin{acronym}
    \acro{DoFs}{Degrees of Freedom}
    \acro{CSR}{Continuum Soft Robot}
    \acro{CRT}{Cosserat Rod Theory}
    \acro{DER}{Discrete Elastic Rod}
    \acro{ODE}{Ordinary Differential Equation}
    \acro{PDE}{Partial Differential Equation}
    \acro{FEM}{Finite Element Method}
    \acro{FT}{Fourier Transform}
    \acro{SFT}{Spatial Fourier Transform}
    \acro{STFT}{Space-Time Fourier Transform}
    \acro{FFT}{Fast Fourier Transform}
    \acro{ZOH}{Zero-Order Hold}
    \acro{FOH}{First-Order Hold}
    \acro{ISP}{Implicit Strain Parameterization}
    \acro{GVS}{Geometrically Variable Strain}
    \acro{PCS}{Piecewise Constant Strain}
    \acro{PLS}{Piecewise Linear Strain}
    \acro{PAS}{Piecewise Affine Strain}
    \acro{PCC}{Piecewise Constant Curvature}
    \acro{PAC}{Piecewise Affine Curvature}
    \acro{SoRoSim}{Soft Robot Simulator}
    \acro{MP}{Matching Pursuit}
    \acro{BPD}{Basis Pursuit Denoising}
    \acro{WLS}{Weighted Least Squares}
    \acro{SALSA}{Split Augmented Lagrangian Shrinkage Algorithm}
    \acro{POD}{Proper Orthogonal Decomposition}
    \acro{SVD}{Singular Value Decomposition}
    \acro{MSE}{Mean Squared Error}
\end{acronym}

\usepackage{algorithm}
\usepackage{algpseudocode}

\newcommand\BibTeX{{\rmfamily B\kern-.05em \textsc{i\kern-.025em b}\kern-.08em
T\kern-.1667em\lower.7ex\hbox{E}\kern-.125emX}}

\def\volumeyear{2026}

\newcommand{\fourier}[1]{\mathcal{F}\left\{#1\right\}}
\newcommand{\rect}[1]{\textnormal{rect}\left(#1\right)}
\newcommand{\sinc}[1]{\textnormal{sinc}\left(#1\right)}
\newcommand{\abs}[1]{\left\lvert#1\right\rvert}
\newcommand{\norm}[1]{\left\lVert#1\right\rVert}

\begin{document}


\title{SoFFT: Spatial Fourier Transform for Modeling Continuum Soft Robots}

\author{Daniele~Caradonna\affilnum{1,2}, Diego~Bianchi\affilnum{1,2}, Franco~Angelini\affilnum{3,4}, and Egidio~Falotico\affilnum{1,2}}

\affiliation{\affilnum{1}The BioRobotics Institute, Scuola Superiore Sant'Anna, Pisa, Italy. \\
\affilnum{2}Department of Excellence in Robotics and AI, Scuola Superiore Sant'Anna, Pisa, Italy. \\
\affilnum{3}Centro di Ricerca ``En\-ri\-co Pi\-ag\-gio'', U\-ni\-ver\-si\-t\`{a} di Pisa, Largo Lucio Lazzarino 1, 56126 Pisa, Italy. \\
\affilnum{4}Dipartimento di Ingegneria dell'Informazione,  U\-ni\-ver\-si\-t\`{a} di Pisa, Largo Lucio Lazzarino 1, 56126 Pisa, Italy.
}

\corrauth{Daniele Caradonna, The BioRobotics Institute, Scuola Superiore Sant'Anna, Pisa, Italy.}

\email{daniele.caradonna@santannapisa.it}

\begin{abstract}
Continuum soft robots, composed of flexible materials, exhibit theoretically infinite degrees of freedom, enabling notable adaptability in unstructured environments. Cosserat Rod Theory has emerged as a prominent framework for modeling these robots efficiently, representing continuum soft robots as time-varying curves, known as backbones. In this work, we propose viewing the robot's backbone as a signal in space and time, applying the Fourier transform to describe its deformation compactly. This approach unifies existing modeling strategies within the Cosserat Rod Theory framework, offering insights into commonly used heuristic methods. Moreover, the Fourier transform enables the development of a data-driven methodology to experimentally capture the robot's deformation. 
The proposed approach is validated through numerical simulations and experiments on a real-world prototype, demonstrating a reduction in the degrees of freedom while preserving the accuracy of the deformation representation.
\end{abstract}


\maketitle

\section{Introduction} \label{sec:introduction}
Soft robotics is a promising area of robotics in which robots are made from soft materials, such as silicones or rubbers \cite{rus2015design}. 
The continuum nature of this type of robot exhibits theoretically infinite \ac{DoFs}, enabling adaptability in unstructured or cluttered environments \cite{del2024growing}. 
However, the infinite \ac{DoFs} of \acp{CSR} make controllers' design arduous.

The soft robotics community proposed numerous controllers, marking two main approach types: model-based \cite{della2023model} and learning-based \cite{george2018control} controllers. The former exploits an accurate robot model to accomplish the task. The latter utilizes experimental or simulated data to train learning algorithms. In both cases, the a priori knowledge provided by the model significantly improves the controllers' performances \cite{falotico2024learning}. In this context, modeling \acp{CSR} became crucial, and researchers tried to find the best trade-off between accuracy and computational efficiency \cite{armanini2023soft}.

In the last decade, the \ac{CRT} \cite{renda2014dynamic, gazzola2018forward} has emerged as one of the most effective compromises, accounting for all possible strain modes, including stretching, shearing, bending, and twisting. Moreover, numerous implementations exist with proven computational efficiency \cite{mathew2022sorosim, naughton2021elastica}.
The main idea of \ac{CRT} for \acp{CSR} is to describe them as time-varying oriented curves, referred to as the backbone. The evolution of the backbone is described by a set of \acp{PDE}, reflecting the infinite \ac{DoFs} inherent in the continuum structures.

\begin{figure}
    \centering
    \includegraphics[width=1.0\linewidth]{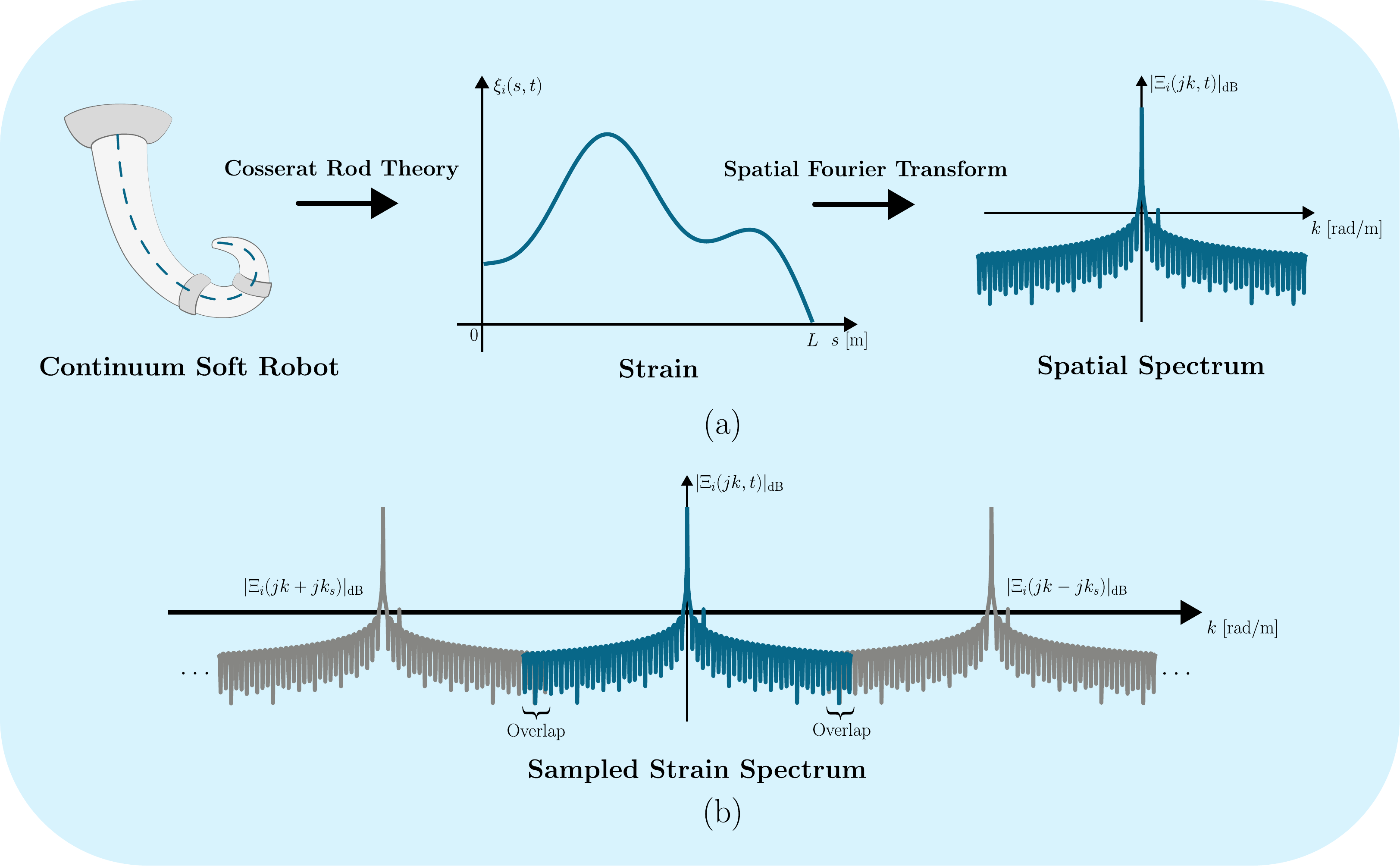}
    \caption{Representative scheme for applying the Fourier transform to \acp{CSR}. (a) The strain field can be extracted from the backbone and analyzed in the spatial frequency domain. (b) The spatial spectrum of a sampled strain field. Due to the unbounded nature of the strain field spectrum, aliasing is inevitable.}
    \label{fig:concept}
\end{figure}

To reduce the number of \ac{DoFs} for simulation and control, various spatial discretization techniques have been proposed \cite{armanini2023soft}. The principal approaches in the literature typically divide the backbone into segments, assuming each segment forms a specific curve (e.g., circumference \cite{webster2010design}, helix \cite{grazioso2019geometrically}). 
Mathematically, this is equivalent to reconstructing a signal in space from a finite set of samples by fitting with a specific set of functional bases \cite{renda2020geometric, boyer2020dynamics}, such as polynomials \cite{della2019control}.

In this work, we take this concept further by treating the backbone as a spatial signal and applying the \ac{FT} \cite{bracewell2007fourier}. This approach provides a compact representation of the robot's deformation and unifies existing modeling strategies within the \ac{CRT} framework.
\color{black}
Rather than simplifying the original \ac{CRT} \acp{PDE}, our framework provides a reinterpretation of these PDEs in terms of their spatial frequency content.
\color{black}
Additionally, it offers a theoretical foundation for heuristic techniques, such as determining the minimum required number of segments.
Furthermore, the \ac{FT} offers a new experimental methodology for selecting optimal modeling methods based on data, balancing accuracy, and computational efficiency.
Finally, we validated the proposed data-driven approach through numerical simulations and real-world experiments.

The paper is organized as follows. In Sec. \ref{sec:background}, we discuss the preliminaries, including a brief description of \ac{CRT} (Sec. \ref{background:cosserat_rod_theory}) and the current state of the art in modeling \acp{CSR} (Sec. \ref{background:soa}). In Secs. \ref{sec:spatial_ft} and \ref{sec:space_time_ft}, we apply the \ac{FT} to the \acp{CSR}. Furthermore, we propose a data-driven methodology in Sec. \ref{sec:spectrum_extraction}. Sections \ref{sec:numerical_validation} and \ref{sec:experimental_validation} validate the approach using both simulated and real-world \acp{CSR}. Finally, Sec. \ref{sec:conclusions} summarizes the contributions and results of this work.

For the sake of readability, the main variables and symbols used are listed in Table \ref{tab:nomenclature}. Additionally, Appendix \ref{sec:appendix} provides the definitions of the Lie Algebra operators and the properties of the \ac{FT}.
\section{Background} \label{sec:background}
    \subsection{Cosserat Rod Theory} \label{background:cosserat_rod_theory}
        The \ac{CRT} associates to the elastic rod of length $L$ an oriented backbone, parameterized by the material curvilinear abscissa $s \in [0, \, L]$. The oriented curve is described by its pose $\bm{g}(s, t) \in SE(3)$ for each cross-section $s$, defined as
\begin{equation} \label{eq:se3_definition}
    \bm{g}(s, t) = \begin{bmatrix}
                \bm{R}(s, t) & \bm{r}(s, t) \\
                \bm{0}^{\top} & 1 
        \end{bmatrix} \, ,
\end{equation}
where $\bm{R}(s, t) \in SO(3)$ is the rotation matrix and $\bm{r}(s, t) \in \mathbb{R}^{3}$ is the position.
The space-evolution of $\bm{g}$ can be uniquely described by the strain field $\bm{\xi}(s, t) = \begin{bmatrix} \bm{\kappa}^{\top} & \bm{\sigma}^{\top} \end{bmatrix}^{\top} \in \mathbb{R}^6$, which is defined as
\begin{equation} \label{eq:strain_field}
    \bm{\xi}(s, t) = \left(\bm{g}^{-1}(s, t) \, \bm{g}'(s, t)\right)^{\vee} \, ,
\end{equation}
where ${\left(\cdot\right)}^{\vee}$ denotes the vee operator \cite[Ch.\ 3]{murray}, $\bm{\kappa} = \begin{bmatrix}
    \kappa_x & \kappa_y & \kappa_z
\end{bmatrix}^{\top} \in \mathbb{R}^{3}$ contains the angular strain modes, such as twisting ($\kappa_x$), and bending ($\kappa_y , \, \kappa_z$), and $\bm{\sigma} = \begin{bmatrix}
    \sigma_x & \sigma_y & \sigma_z
\end{bmatrix}^{\top} \in \mathbb{R}^{3}$ represents the linear strain modes, such as stretching ($\sigma_x$), and shear ($\sigma_y, \, \sigma_z$).

The time evolution of $\bm{g}(s, t)$ can be described with the velocity field $\bm{\eta}(s, t) \in \mathbb{R}^6$, defined as
\begin{equation} \label{eq:velocity_field}
    \bm{\eta}(s, t) = \left(\bm{g}^{-1}(s, t) \, \dot{\bm{g}}(s, t)\right)^{\vee} \, .
\end{equation}
Moreover, the mixed derivative equality allows us to write a relationship between the two fields, i.e.,
\begin{equation} \label{eq:mixed_derivative}
    \bm{\eta}'(s, t) = \dot{\bm{\xi}}(s, t) - \textnormal{ad}_{\bm{\xi}} \, \bm{\eta}(s, t) \, . 
\end{equation}

To model the statics and dynamics of a Cosserat rod, both internal and external wrenches must be defined. The internal wrench consists of the passive internal wrench $\bm{\mathcal{F}}_i(s, t) \in \mathbb{R}^6$, which accounts for the mechanical impedance of the rod (i.e., its elasticity and damping), and the active internal wrench $\bm{\mathcal{F}}_a(s, t) \in \mathbb{R}^6$, which represents the distributed actuation applied to the \acp{CSR}.
\begin{equation} \label{eq:internal_passive}
    \bm{\mathcal{F}}_i (s, t) = \bm{\Sigma}(s) \left(\bm{\xi} - \bm{\xi}^*\right) + \bm{\Psi}(s) \, \dot{\bm{\xi}} \, ,
\end{equation}
\begin{equation} \label{eq:internal_active}
    \bm{\mathcal{F}}_a(s, t) = \bm{B}_{\bm{\tau}}(\bm{\xi}, s) \, \bm{\tau}(t) \, .
\end{equation}
In \eqref{eq:internal_passive}, $\bm{\Sigma}(s) = \textnormal{diag} \left(G J_x, E J_y, E J_z, E A, G A, G A \right) \in \mathbb{R}^{6 \times 6}$ is the stiffness matrix, $\bm{\Psi} = \beta  \, \textnormal{diag} \left(J_x, 3 J_y, 3 J_z, 3 A, A, A \right) \in \mathbb{R}^{6 \times 6}$ the damping matrix, where $A(s)$ and $J_i(s)$ are the cross-section's area and the second moment of area, respectively. 
Furthermore, $E(s)$ is the Young modulus, $G(s)$ is the shear modulus, $\beta(s)$ is the damping coefficient, and $\bm{\xi}^{*}(s) \in \mathbb{R}^{6}$ is the stress-free strain field.

Concerning \eqref{eq:internal_active}, $\bm{B}_{\bm{\tau}} \in \mathbb{R}^{6 \times n_a}$ is the actuation matrix \cite{renda2017screw}, $\bm{\tau} \in \mathbb{R}^{n_a}$ is the actuators' magnitude vector, and $n_a$ is the number of the actuators.
The external wrench $\bm{\mathcal{F}}_e(s, t) \in \mathbb{R}^6$ collects the distributed external forces applied to the robot, such as gravity or contact forces; the contribution of gravity is detailed in the following
\begin{equation} \label{eq:ext_gravity}
    \bm{\mathcal{F}}_{e, \, g}(s, t) = \bm{\mathcal{M}}(s) \, \textnormal{Ad}^{-1}_{\bm{g}} \, \bm{\mathcal{G}} \, ,
\end{equation}
where $\bm{\mathcal{G}} \, \textcolor{black}{= \begin{bmatrix} 0 & 0 & 0 & a_{\textnormal{g}} & 0 & 0  \end{bmatrix}^{\top}} \in \mathbb{R}^{6}$ is the gravity acceleration twist, \textcolor{black}{$a_{\textnormal{g}} \in \mathbb{R}$ the gravitational acceleration,} $\bm{\mathcal{M}} = \rho \, \textnormal{diag}\left(J_x, J_y, J_z, A, A, A\right)$ is the cross-section's inertia matrix, and $\rho(s)$ is the cross-section's density.

The statics of a Cosserat Rod can be written as
\begin{equation} \label{eq:crt_statics}
    \left(\bm{\mathcal{F}}_i - \bm{\mathcal{F}}_a\right)' + \textnormal{ad}_{\bm{\xi}}^{*} \left(\bm{\mathcal{F}}_i - \bm{\mathcal{F}}_a\right) + \bm{\mathcal{F}}_e = \bm{0} \, ,
\end{equation}
where, in \eqref{eq:internal_passive}, the damping contribution is neglected due to the static regime.

Finally, the dynamics of the Cosserat rod can be derived using the Poincaré principle \textcolor{black}{\cite{boyer2017poincare}}, yielding 
\begin{equation} \label{eq:crt_dynamics}
    \begin{split}
        \bm{\mathcal{M}} \dot{\bm{\eta}} &+ \textnormal{ad}^*_{\bm{\eta}} \left(\bm{\mathcal{M}} \bm{\eta}\right) = \\
        \left(\bm{\mathcal{F}}_i - \bm{\mathcal{F}}_a \right)' &+ \textnormal{ad}^*_{\bm{\xi}} \left(\bm{\mathcal{F}}_i - \bm{\mathcal{F}}_a \right) + \bm{\mathcal{F}}_{e} \, .
    \end{split}
\end{equation}
    \begin{table}[h!]
\small
\centering
\caption{Nomenclature. \textcolor{black}{UoM stands for Unit of Measurement.}}
\label{tab:nomenclature}
    \begin{tabular}{lll}
    \toprule
    Name & Symbol & \textcolor{black}{UoM} \\
    \midrule
    Length & $L \in \mathbb{R}^{+}$ & \si{\meter} \\
    Material Abscissa & $s \in [0, \, L]$ & \si{\meter} \\
    Time & $t \in [0, \, +\infty)$ & \si{\second} \\
    Cross-Section's Area & $A(s) \in \mathbb{R}^{+}$ & \si{\meter^{2}} \\
    Cross-Section's Density & $\rho(s) \in \mathbb{R}^{+}$ & \si{\kilogram / \meter^{2}} \\
    Young Modulus & $E(s) \in \mathbb{R}^{+}$ & \si{\pascal} \\
    Damping Coefficient & $\beta(s) \in \mathbb{R}^{+}$ & \si{\pascal \cdot \second} \\
    Wavelength & $\lambda \in \mathbb{R}$ & \si{\meter} \\
    Wavenumber & $\nu = \lambda^{-1}$  & \si{\meter^{-1}}\\
    Angular Wavenumber & $k = 2 \pi \nu$ & \si{\radian / \meter} \\
    Period & $T \in \mathbb{R}$ & \si{\second} \\
    Frequency & $f = T^{-1}$ & \si{\hertz} \\
    Angular Frequency & $\omega = 2 \pi f$ & \si{\radian / \second} \\
    \midrule
    Name & Symbol & \ac{FT} \\
    \midrule
    Strain Field & $\bm{\xi}(s, t) \in \mathbb{R}^6$ & $\bm{\Xi}(jk, \, j \omega)$ \\
    Stress-Free Strain & $\bm{\xi}^{*}(s) \in \mathbb{R}^6$ & $\bm{\Xi}^{*}(jk)$  \\
    Velocity Field & $\bm{\eta}(s, t) \in \mathbb{R}^6$ & $\bm{\mathcal{H}}(jk, \, j \omega)$ \\
    Passive Internal Wrench & $\bm{\mathcal{F}}_i(s, t) \in \mathbb{R}^6$ & $\bm{F}_i(jk, \, j \omega)$ \\
    Active Internal Wrench & $\bm{\mathcal{F}}_a(s, t) \in \mathbb{R}^6$ & $\bm{F}_a(jk, \, j \omega)$ \\
    External Wrench & $\bm{\mathcal{F}}_e(s, t) \in \mathbb{R}^6$ & $\bm{F}_e(jk, \, j \omega)$ \\
    Virtual Joint Vectors & $\bm{q}(t) \in \mathbb{R}^{n_q}$ & $\bm{Q}(j \omega)$ \\
    Actuators' Magnitude & $\bm{\tau}(t) \in \mathbb{R}^{n_a}$ & $\bm{T}(j \omega)$ \\
    Inertia Matrix & $\bm{\mathcal{M}}(s) \in \mathbb{R}^{6}$ & $\bm{M}(jk)$ \\
    Stiffness Matrix & $\bm{\Sigma}(s) \in \mathbb{R}^{6 \times 6}$ & $\bm{\Sigma}(jk)$ \\
    Damping Matrix & $\bm{\Psi}(s) \in \mathbb{R}^{6 \times 6}$ & $\bm{\Psi}(jk)$ \\
    Actuation Matrix & $\bm{B}_{\bm{\tau}}\left(\bm{\xi}, \, s\right) \in \mathbb{R}^{6 \times n_a}$ & $\bm{B}_{\bm{\tau}}\left(\bm{\Xi}, \, jk\right)$ \\
    Basis Matrix & $\bm{B}_{\bm{q}}(s) \in \mathbb{R}^{6 \times {n_q}}$ & $\bm{B}_{\bm{q}}(jk)$ \\
    \bottomrule
    \end{tabular}
\end{table}

    \subsection{State of the Art} \label{background:soa}
        The dynamics of a Cosserat rod \eqref{eq:crt_dynamics} is a set of \acp{PDE}, whose solution lies in a functional space, i.e., an infinite-dimension space. However, many approaches have been proposed in the literature to efficiently describe the robot with a finite number of \ac{DoFs}, trying to find the best trade-off between accuracy and computational efficiency.

One of the most common modeling approaches is the \ac{PCC} \cite{webster2010design}. The core idea is to divide the backbone into segments and assume them as circumference arcs. This approximation fits very well for the slender shape of the \acp{CSR}. Furthermore, it provides the dynamics in the classical Lagrangian form \cite{siciliano}, allowing the transfer of many controllers from the rigid robot literature \cite{della2018dynamic}. 
In addition, the \ac{PCC} approach accounts for both bending and elongation modes, \textcolor{black}{effectively describing the deformations induced by actuators routed parallel to the backbone, under the assumption of negligible external force contributions.}

To include the twisting mode, \cite{grazioso2019geometrically} proposed to fit the pieces with a helix (i.e., constant curvature and torsion), enabling a general curve in the 3D space.

However, the approximation of unshearable rods is unsuitable for interaction tasks where shear deformation is significant, particularly in cases involving contact with irregular objects or the environment. To expand the previous models, \cite{renda2018discrete} proposed the \ac{PCS} approach, which consists of considering the strain field constant along the length of the single piece. 

Differently, \cite{della2020soft} considers the curvature an affine function instead of constant for each single segment.
With only two \ac{DoFs}, this model exhibits nonlinear phenomena such as snap effect \cite{armanini2017elastica, caradonna2024model}. A further work \cite{stella2023piecewise} proposed the \ac{PAC}, in which each piece is described by an affine function in $s$.

Following this methodology, \cite{li2023piecewise} presented the \ac{PLS}, in which the strain field is computed by linear interpolation of samples. In addition, \ac{PLS} is applied in the case of interactions, exploiting excellent performances w.r.t. the other modeling methodologies \cite{xun2024cosserat}.
In general, we can call \ac{PAS} all the approaches that use the linear approximation of each segment of the strain field. 

Differently from the piecewise methods, in \cite{renda2020geometric, boyer2020dynamics} the \ac{GVS} is proposed, in which the strain field is assumed as generated by a truncated functional basis of space-dependent vectors \cite{armanini2023soft}. It is worth highlighting that, with a polynomial basis function, the \ac{GVS} approach coincides with the affine strain and curvature approximation. However, the \ac{GVS} approach allows using other basis functions, such as trigonometric and Gaussian.

Furthermore, utilizing the Magnus expansion \cite[Chap. IV.7]{hairergeometric} and Zanna's collocation method \cite{zanna1999collocation}, the Authors provided an efficient and recursive algorithm to compute the Jacobian of the \ac{CSR}. 
The primary advantage of this approach is that, regardless of the functional basis chosen, the classic Lagrangian form \cite{siciliano} of the dynamics can be found.
This significantly eases the transfer of classical controllers from rigid to soft robotics.

 Following this approach, the Authors proposed a strain-dependent functional basis called \ac{ISP} \cite{renda2024dynamics}. In this approach, the basis functions are derived from the robot's statics, achieving a minimal number of \ac{DoFs} corresponding to the strain modes excited by the actuators’ routing and external forces. However, under dynamic regimes, further strain modes might be excited, requiring a user-specified extended basis.

In \cite{mathew2024reduced}, the Authors present a comprehensive explanation of the \ac{GVS} framework, covering every possible type of functional basis. 
Additionally, it has been incorporated into \ac{SoRoSim} \cite{mathew2022sorosim}, which allows for rapid simulation of soft robots, even when they involve numerous amounts of \ac{DoFs}.

Since the choice of basis is critical, the research community has explored methods to optimize the number of bases and reduce the system's order.
For instance, in \cite{pustina2024nonlinear}, the Authors introduced eigenmanifolds to perform modal analysis for \acp{CSR}. This approach facilitates the evaluation of \ac{PCC} models with an increasing number of segments by employing similarity metrics compared to high-fidelity models such as the Finite Element Method.

Finally, \cite{alkayas2024soft} presents a data-driven reduction method based on \ac{POD}. The key concept involves applying \ac{SVD} to the strain data to identify the least significant singular values, enabling truncation without significant loss of accuracy. 
First, a \ac{GVS} digital twin with a high number of \ac{DoFs} is derived by fitting the experimental data. Then, after simulating the fitted \ac{GVS} model, a reduction in the number of \ac{DoFs} is performed using the \ac{POD} method.
\color{black}
Additionally, the Authors extended this approach by incorporating nonlinear strain bases in \cite{alkayas2025structure} employing auto-encoders.
\color{black}

In this context, we propose a novel approach that treats the backbone of the continuum robot as a space-time varying signal and analyzes it using the \ac{FT}. This framework unifies existing methods by interpreting them as reconstructors based on discrete strain samples. Moreover, this perspective provides theoretical justification for various heuristic methods, such as the minimum number of segments required in piecewise approaches.

Additionally, by applying the \ac{FT} to experimental data, we establish a data-driven methodology with a strong mathematical foundation for identifying the optimal functional bases from a predefined signal dictionary. This approach relies solely on the geometric properties of the robot and its actuators. 
The method detects the functional bases in the dynamic situation and extends the static strain analysis of the \ac{ISP} method.

\color{black}
Notably, the goal of employing the \ac{FT} is not to replace the \ac{CRT} \acp{PDE}, but rather to exploit the spectral information to identify the optimal basis within the \ac{GVS} modeling framework.
\color{black}

\color{black}
\subsubsection{Comparison with \ac{POD} \cite{alkayas2024soft}}
The \ac{POD} method extracts optimal spatial bases by applying \ac{SVD} to a snapshot matrix of strain data, enabling energy-based truncation via the singular values. Similarly, our approach uses the \ac{FFT} to directly identify optimal bases from experimental data, providing a unified and theoretically grounded framework for spatial discretization techniques.

Both methods achieve compact representations, but they differ in how strain modes are treated. 
\ac{POD} derives a single coupled basis for all modes, whereas our approach treats each mode independently—avoiding scaling issues and allowing mode-specific truncation. 
In this respect, our method can be suboptimal, as \ac{POD} can capture all modes efficiently with few \ac{DoFs}.

Finally, the \ac{FFT} is computationally more efficient, thanks to its logarithmic complexity, making it better suited for potential real-time applications. Overall, both methods are primarily used offline, meaning that this computational advantage is currently limited but could be leveraged in future real-time implementations.
\color{black}
\section{Spatial Fourier Transform} \label{sec:spatial_ft}
    As seen in Sec. \ref{background:cosserat_rod_theory}, the backbone of a \ac{CSR} can be completely described by the strain field $\bm{\xi}: [0, \, L] \times [0, \, +\infty) \rightarrow \mathbb{R}^6$. The main idea of this work is to consider it as a signal in space and time, applying the \ac{FT} to analyze the robot’s deformation.

To apply it, we must adapt the domain of the strain field. \textcolor{black}{Hence, we assume the domain of $s$ extends to $\mathbb{R}$, with $\bm{\xi}: \mathbb{R} \times [0, +\infty) \rightarrow \mathbb{R}^{6}$, and the strain field is null for all $s \notin [0, L]$.} This assumption is equivalent to applying a spatial window over the interval $[0, \, L]$.

To formalize this, let $\bm{\Phi}(s): \mathbb{R} \rightarrow \mathbb{R}^6$ be a function such that $\bm{\xi}(s) = \bm{\Phi}(s) \, \forall s \in [0, \, L]$. 
The previous assumption is equivalent to stating that $\bm{\xi}(s, t) = \bm{\Phi}(s) \cdot \Pi_L(s)$, where $\Pi_L$ is the window function in the range $[0, \, L]$ \textcolor{black}{defined in \eqref{eq:length_space_window}}.
\begin{figure}
    \centering
    \includegraphics[width=1.0\linewidth]{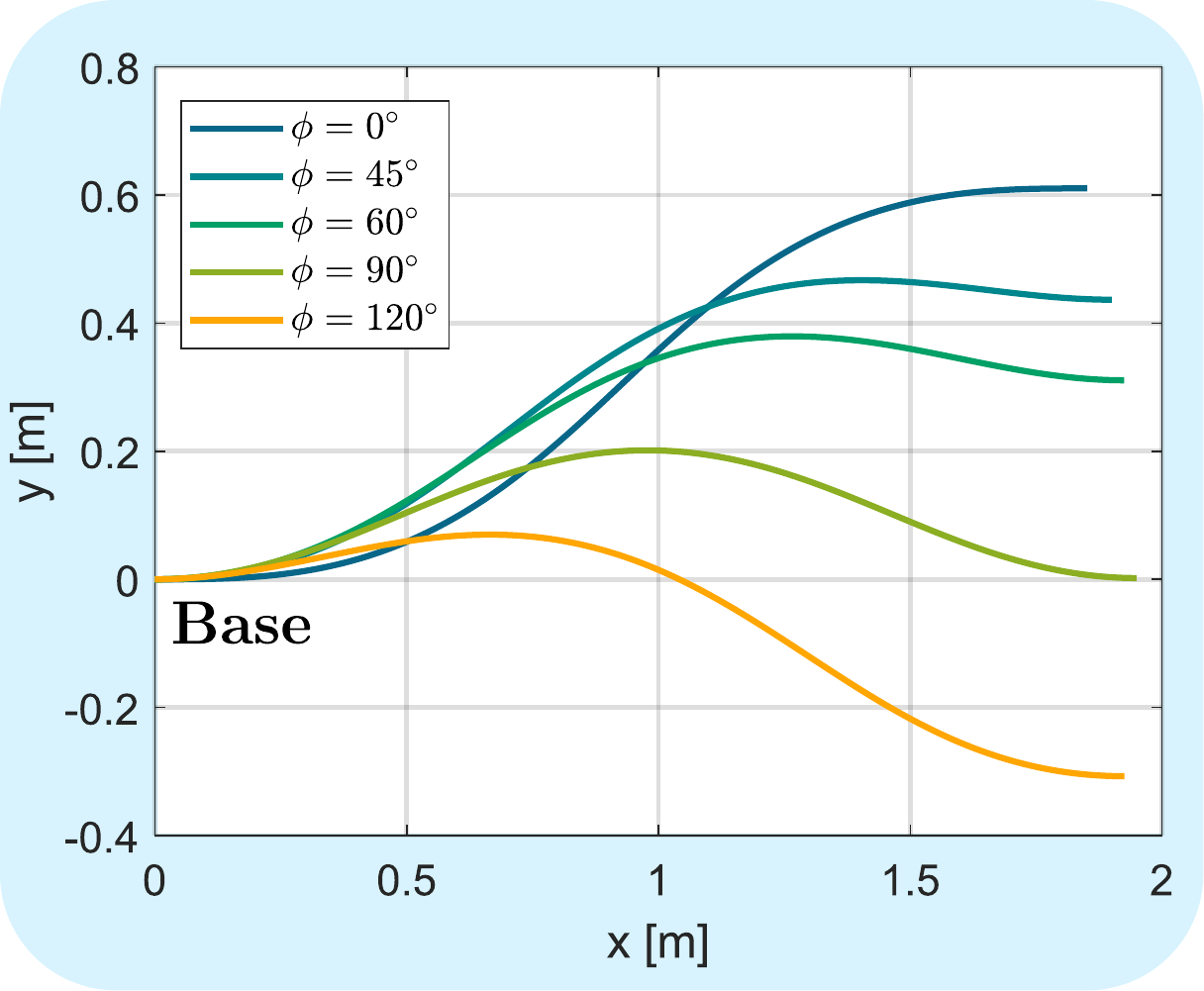}
    \caption{Impact of the phase in a planar rod with a sinusoidal curvature function. Varying the phase, the deformation is distributed differently along the rod.}
    \label{fig:space_shift_example}
\end{figure}
%
\begin{figure*}
    \centering
    \includegraphics[width=1.0\linewidth]{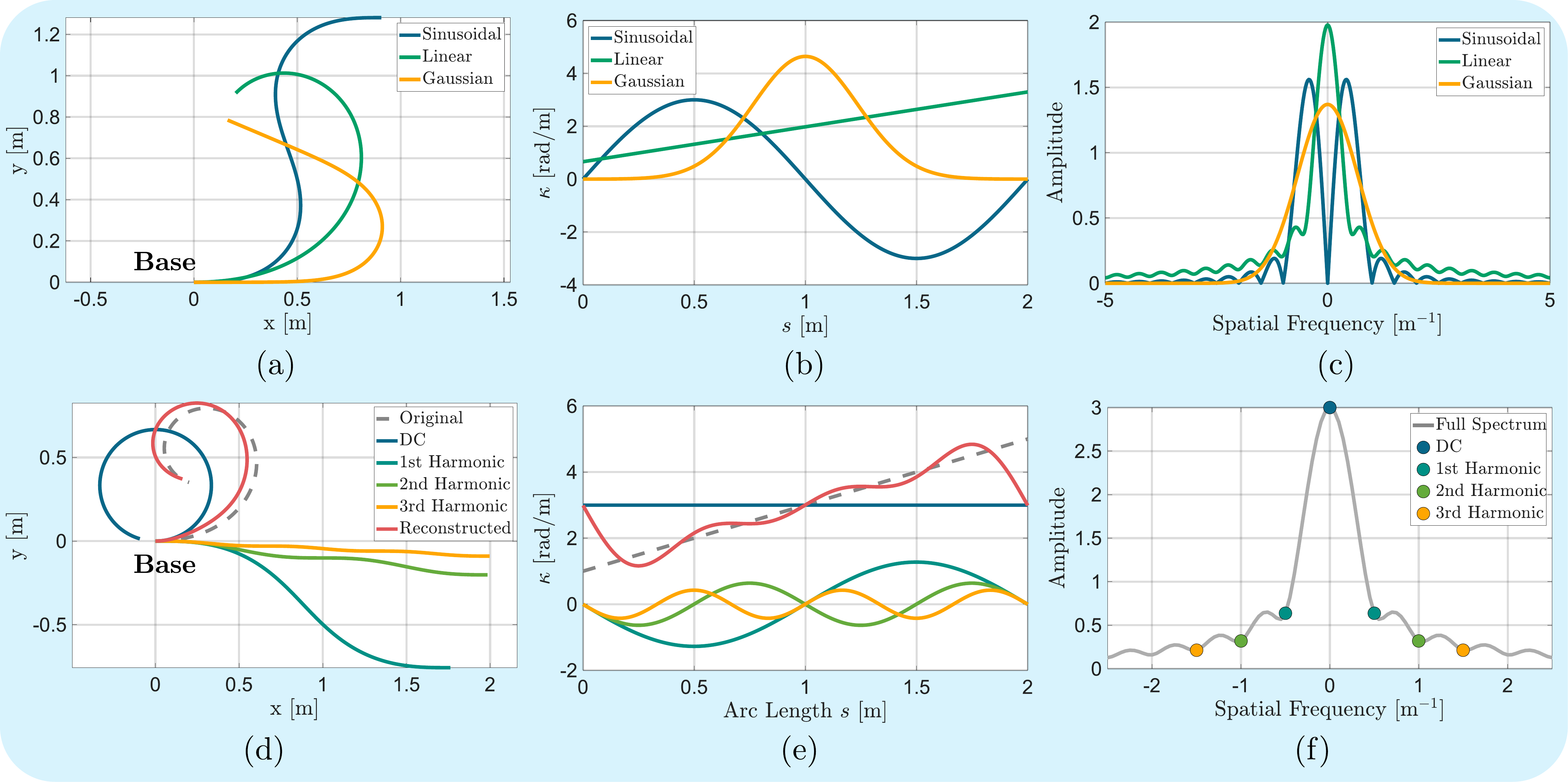}
    \caption{\textcolor{black}{Mechanical interpretation of the Spatial Fourier Transform (SFT) applied to the backbone of a \ac{CSR}. Panels (a–c) show three different curvature profiles (sinusoidal, linear, and Gaussian), along with their corresponding robot shapes and \ac{SFT} representations. Panels (d–f) illustrate the corresponding shapes associated with each spatial harmonic.}}
    \label{fig:mechanical_reality}
\end{figure*}
\subsection{Continous \ac{SFT}} \label{spatial_ft:csft}
The Continous \ac{SFT} of the strain field $\bm{\xi}$ is defined as
\begin{equation} \label{eq:continous_spatial_fourier_transform}
    \bm{\Xi}\left(jk, \, t\right) = \fourier{\bm{\xi}(s, t)} = \int_{0}^{L} \bm{\xi}(s, \, t) \, e^{-jks} \, \textnormal{d} s \, ,
\end{equation}
where $k = 2 \pi \nu \in \mathbb{R}$ is the angular wavenumber, $\nu \in \mathbb{R}$ is the wavenumber (i.e., spatial frequency), and $j = \sqrt{-1}$ the imaginary unity. In this case, the signal $\bm{\xi}(s, t)$ is a signal limited in space ($s \in [0, L]$) and aperiodic. Hence, the spatial spectrum 
$\bm{\Xi}(j k, t)$ will be infinite, composed of all the spatial frequencies, as shown in Fig. \ref{fig:concept}(a). 
This consideration is coherent with the infinite \ac{DoFs} of a \ac{CSR}, considering the strain field generated by an infinite-dimensional trigonometric functional basis.
\color{black}
In special cases, such as constant or sinusoidal strain, the \ac{SFT} displays pronounced harmonics, suggesting that an infinite number of harmonics is not required to represent the strain with negligible error.
\color{black}

\color{black}
The \ac{SFT} exists if the strain field satisfies the conditions for the existence of \ac{FT}. Specifically, $\bm{\xi}(s,t)$ must be absolutely integrable, i.e., $\int_{0}^{L} \left|\bm{\xi}\right| < + \infty$, and it must contain only a finite number of discontinuities.
It is worth noting that the strain field can generally be discontinuous in space, as assumed in piecewise approaches. In such cases, the number of discontinuities remains finite, corresponding to the number of pieces.
\color{black}

Moreover, the \ac{SFT} of the strain field $\bm{\Xi}(jk) \in \mathbb{C}^{6}$ can be analyzed by examining its magnitude and phase. 
The magnitude $\abs{\bm{\Xi}(jk)}$ quantifies the extent of deformation of the slender body, whereas the phase $\angle\bm{\Xi}(jk)$ indicates how the deformation is distributed along the rod, impacting with the rod's shape.
Consider, for instance, a planar backbone characterized by a sinusoidal curvature $\kappa = \sin\left(\pi s + \phi \right) \in \mathbb{R}$.
As illustrated in Fig. \ref{fig:space_shift_example}, this sinusoidal curvature generates a backbone shape that changes as the phase $\phi$ varies. Although the magnitude of the curvature remains constant, the strain is differently distributed along the rod, leading to variations in the overall shape of the backbone.

\color{black}
Several interpretations of the \ac{SFT} can be highlighted in relation to the robot's mechanical reality.
To illustrate this, Fig.~\ref{fig:mechanical_reality}(a-c) presents a 2D rod with curvature defined by three distinct functions: (i) sinusoidal, (ii) linear, and (iii) Gaussian. The parameters are normalized such that each signal contains the same energy. As the curvature profile varies, both the \ac{SFT} and the resulting robot shape change accordingly. This relationship can be exploited practically: if the spatial spectrum can be measured from experimental data, it becomes possible to identify the analytical function that best fits the physical curvature profile.

Another key insight extracted from the \ac{SFT} is the physical contribution of each spatial harmonic to the overall shape of the robot. Figure~\ref{fig:mechanical_reality}(d-f) depicts the same rod with a linear curvature profile alongside its \ac{SFT}. Specifically, Fig.~\ref{fig:mechanical_reality}(d) visualizes the kinematic contribution of each spatial harmonic up to the third order. While the DC component represents the constant curvature contribution that defines the primary base arc, the higher harmonics introduce shapes characterized by more localized curvature gradients, mechanically corresponding to higher-order bending modes.
\color{black}

\color{black}
It is worth noting that the \ac{SFT} can also be applied to position-based formulations of \ac{CRT}, such as \ac{DER} \cite{gazzola2018forward}. However, the strain-based formulation is generally preferred for control purposes, as it provides a more compact representation of the dynamics with fewer \ac{DoFs} and yields Lagrangian equations of motion \cite{boyer2020dynamics}.
\color{black}

\subsection{Statics of a Cosserat Rod in the Spatial Frequency Domain}
We can exploit the properties of \ac{FT} \eqref{eq:fourier_definition}-\eqref{eq:fourier_conv} in the case of the Spatial Spectrum $\bm{\Xi}(jk)$, allowing us to rewrite \eqref{eq:crt_statics} in the spatial frequency domain, such as
\begin{equation} \label{eq:static_ft}
    jk \, \left(\bm{F}_i - \bm{F}_a\right) + \textnormal{ad}_{\bm{\Xi}}^{*} \ast \left(\bm{F}_i - \bm{F}_a\right) \textcolor{black}{+ \bm{F_e}} = \bm{0} \, ,
\end{equation}
where $\bm{F}_i(jk) = \fourier{\bm{\mathcal{F}}_i(s)}$, $\bm{F}_a(jk) = \fourier{\bm{\mathcal{F}}_a(s)}$, \textcolor{black}{$\bm{F}_e(jk) = \fourier{\bm{\mathcal{F}}_e(s)}$}, and $\ast$ is the convolution in the spatial frequency domain.

In the case of internal passive wrench $\bm{\mathcal{F}}_i(s)$, the \ac{FT} can be expressed as
\begin{equation} \label{eq:internal_passive_static_ft}
    \bm{F}_i(j k) = \bm{\Sigma}(jk) \ast \left(\bm{\Xi}(jk) - \bm{\Xi}^{*}(jk) \right) \, ,
\end{equation}
where $\bm{\Sigma}(jk) = \fourier{\bm{\Sigma}(s)}$ is the \ac{SFT} of the stiffness matrix. It is worth highlighting that the stiffness matrix $\bm{\Sigma}(s)$ acts as a convolutional filter applied to the strain field in the spatial frequency domain.

In the case of internal active wrench $\bm{\mathcal{F}}_a(s)$, the \ac{SFT} can be expressed as
\begin{equation}
    \bm{F}_a(j k) = \bm{B}_{\bm{\tau}}(\bm{\Xi}, jk) \, \bar{\bm{\tau}} \, ,
\end{equation}
where $\bm{B}_{\bm{\tau}}(\bm{\Xi}, jk) = \fourier{\bm{B}_{\bm{\tau}}(\bm{\xi}, s)}$ is the \ac{SFT} of the actuation matrix, and $\bar{\bm{\tau}} \in \mathbb{R}^{n_a}$ a generic constant input.

\subsection{Discrete \ac{SFT} and Sampled Strain Field}
For each time instant, let us assume to have $N$ samples of the strain field equally spaced along the rod $\bm{\xi}(n \lambda_s, t)$, where $n \in [0, N - 1]$, and $\lambda_s = L / N$ is the sampling wavelength. These samples can be directly measured using shape sensors \cite{floris2021fiber, feliu2025advancing}, motion capture systems \cite{field2009motion}, or estimated using shape estimation algorithms (e.g., \cite{lilge2022continuum, feliu2025actuation}).
In the discrete case, we can define Discrete \ac{SFT} of the strain field as
\begin{equation} \label{eq:discrete_spatial_fourier_transform}
    \bm{\Xi}(jk, t) = \sum_{n = 0}^{N - 1} \bm{\xi}(n \lambda_s, t) \, e^{-jk n \lambda_s} \, . 
\end{equation}
Moreover, the sampled signal $\bm{\xi}(n \lambda_s, t)$ in the spatial spectrum can be expressed as
\begin{equation} \label{eq:repetition_spectrum}
    \bm{\Xi}_{\textnormal{d}}(jk, t) = \frac{1}{\lambda_s} \sum_{n = - \infty}^{+ \infty} \, \bm{\Xi}\left(jk - j n k_s, \, t\right) \, ,
\end{equation}
where $k_s = \frac{2 \pi}{\lambda_s}$ denotes the sampling angular wavenumber. Coherently with time-varying signals, the spatial spectrum of the discretized strain field consists of an infinite series of replicas of the \ac{SFT}, each shifted by a multiple of $k_s$. Since the spatial spectrum of $\bm{\xi}$ is unlimited, the aliasing effect is inevitable, as illustrated in Fig. \ref{fig:concept}(b). 
 To address this, we can select the last most significant component $k_{\textnormal{max}} \in \mathbb{R}$ and apply the Nyquist-Shannon theorem \cite{shannon1949communication}, resulting in $k_s > 2 \, k_{\textnormal{max}}$.

In the context of modeling \acp{CSR}, it can be more meaningful to express the previous relation in terms of the wavelength, such as
\begin{equation} \label{eq:nyquist_wavelength}
    \lambda_{s} < \frac{\lambda_{\textnormal{max}}}{2} \rightarrow N > \frac{2}{\lambda_{\textnormal{max}}} L \, ,
\end{equation}
where $\lambda_{\textnormal{max}} = \frac{2 \pi}{k_{\textnormal{max}}}$. Eq. \eqref{eq:nyquist_wavelength} provides useful information about the minimum number of pieces a \ac{CSR}, assuming that most of the signal's energy is concentrated up to $k_{\textnormal{max}}$.
This result supports the intuitive and heuristic understanding that more segments are needed for longer rods, as $\lambda_s = L / N$.

\subsection{Energy-based Criterion for Truncation} \label{spatial_ft:truncation}
To determine $k_{\textnormal{max}}$, it is possible to compute the quantity of the signal energy contained up the component $k_{\textnormal{max}}$ and compare it to the total energy of the signal.
Recalling \eqref{eq:discrete_spatial_fourier_transform} and the Parseval identity \eqref{eq:fourier_parseval} in the discrete domain \textcolor{black}{\eqref{eq:discrete_parseval}}, the truncation index $E_{\textnormal{tr}, \xi_i}$ can be defined as
\begin{equation} \label{eq:discrete_truncation_criterion}
    E_{\textnormal{tr}, \xi_i}\left(k_{\textnormal{max}}\right) = \frac{N}{N_{\textnormal{max}}} \left(\frac{\sum_{n = 0}^{N_{\textnormal{max}} - 1} |\Xi_{i}(n)|^{2}}{\sum_{n = 0}^{N - 1} |\Xi_{i}(n)|^{2}}\right) \, ,
\end{equation}
where $N_{\textnormal{max}} < N$ is the number of the sample associated with the angular wavenumber $k_{\textnormal{max}}$, \textcolor{black}{$\xi_i \in \mathbb{R}$ is the $i$-th element of the strain field $\bm{\xi}$, and $\Xi_i = \fourier{\xi_i}$.} 

The truncation index represents the ratio of the energy accumulated up to $k_\textnormal{max}$,  
\textcolor{black}{\(\frac{1}{N_{\textnormal{max}}} \sum_{n = 0}^{N_{\textnormal{max}} - 1} |\Xi_i(n)|^2\)},  
to the total signal energy,  
\textcolor{black}{\(\frac{1}{N} \sum_{n = 0}^{N - 1} |\Xi_i(n)|^2\)}, giving a measure of the truncation accuracy.

Moreover, the truncation index can assume an interesting physical interpretation. By rewriting \eqref{eq:discrete_truncation_criterion} in terms of internal passive wrench \eqref{eq:internal_passive_static_ft}, the truncation index represents the ratio between the deformation energy stored up to $k_{\textnormal{max}}$ and the total deformation energy of the \ac{CSR}. 
Furthermore, the stiffness matrix differentiates between strain modes, leveraging the robot's geometric and physical characteristics.

\subsection{Interpretation of Spatial Discretization Techniques in the Spatial Frequency Domain}
\begin{figure*}
    \centering
    \includegraphics[width=1.0\linewidth]{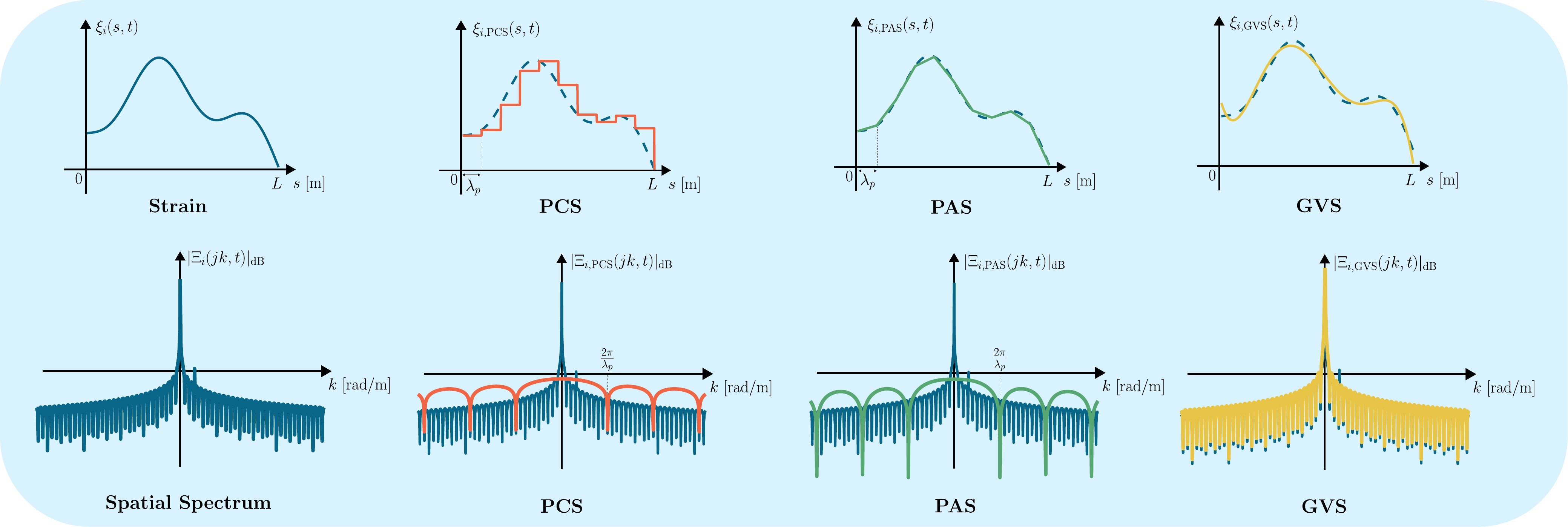}
    \caption{Comparison of the different spatial discretization methodologies. By treating the strain field as a signal, existing modeling approaches can be interpreted as reconstructors.}
    \label{fig:spectra}
\end{figure*}
From the samples $\bm{\xi}(n \lambda_s)$, the spatial discretization techniques exposed in the Sec. \ref{background:soa} can be seen as signal reconstructors.

The \ac{PCS} method reconstructs the sampled strain, assuming constant strain along the single piece. Assuming $n_p$ pieces with the same length $\lambda_p = L / n_p$, the strain field can be written as
\begin{equation} \label{eq:pcs_strain}
    \xi_i(s, t) = \sum_{h = 0}^{n_{p} - 1} q_h(t) \, \Pi_h\left(s\right) \, ,
\end{equation}
where $\Pi_h\left(s\right)$ \textcolor{black}{is defined in \eqref{eq:piece_window}}, and $q_h \in \mathbb{R}$ is the constant value assumed in the $h$-th piece by $\xi_i$. We can operate the \ac{SFT} of \eqref{eq:pcs_strain}, resulting in 
\begin{equation}
    \Xi_{i, \textnormal{PCS}}(jk, t) = \sum_{h = 0}^{n_p - 1} q_i(t) \, \Pi_h\left(jk\right) \, ,
\end{equation}
where $\Pi_h\left(jk\right)$ \textcolor{black}{is described in \eqref{eq:sft_piece_window}}.
If the coefficients $q_h$ correspond to the $\xi_i(h \lambda_p)$, the \ac{PCS} approach can be interpreted as a spatial \ac{ZOH} of the discrete signal $\bm{\xi}$. The reconstructed signal through \ac{PCS} can be expressed as $\bm{\Xi}_{\textnormal{PCS}}(jk) = \bm{H}_0 (jk)  \, \bm{\Xi}_d(jk)$, where \textcolor{black}{$\bm{H}_0(jk)$ is stated in \eqref{eq:zoh_pcs}.}

Similar to the time \ac{ZOH}, the \ac{PCS} approach provides a shift in space of $\lambda_p / 2$, as evident in Fig. \ref{fig:spectra}. Furthermore, for any multiple of $2 \pi / \lambda_p$, the magnitude of $\bm{H}_0(jk)$ falls to 0. Note that these considerations are valid also for the \ac{PCC} case, due to the generality of \ac{PCS} method.

Concerning the \ac{PAS} method, we can write the reconstructed strain as
\begin{equation} \label{eq:pas_strain}
    \xi_i(s, t) = \sum_{h = 0}^{n_p - 1} \left(q_{h, 0}(t) + q_{h, 1}(t) s \right) \Pi_h\left(s\right) ,
\end{equation}
where $q_{h, \, 0}, \, q_{h, \, 1} \in \mathbb{R}$ are the time-varying coefficients of the $i$-th element of the strain field.
In the case of \ac{PLS}, it coincides with the \ac{FOH} reconstructor in space. The reconstructed spectrum can be written as $\bm{\Xi}_{\textnormal{PLS}}(jk) = \bm{H}_1(jk) \, \bm{\Xi}_d(jk)$, where \textcolor{black}{$\bm{H}_1(jk)$ is provided in \eqref{eq:foh_pas}.}

Differently from the piecewise methods, the \ac{GVS} can be seen as a fitting of the samples $\bm{\xi}(n \lambda_s)$ with specific functional bases, such as
\begin{equation} \label{eq:gvs_strain}
    \bm{\xi}(s, t) = \bm{B}_{\bm{q}}(s) \, \bm{q}(t) + \bm{\xi}^{*}(s) \, ,
\end{equation}
where $\bm{B}_{\bm{q}}(s) \in \mathbb{R}^{6 \times n_q}$ is the functional basis matrix and $\bm{q} \in \mathbb{R}^{n_q}$ is the time-dependent coefficients, considered as a vector of virtual joint variables.

The \ac{SFT} of the \ac{GVS} can be expressed as
\begin{equation} \label{eq:gvs_sft}
    \bm{\Xi}_{\textnormal{GVS}}(jk, t) = \left(\bm{B}_{\bm{q}}\left(jk\right) \, \bm{q}(t) + \bm{\Xi}^{*} \right) \ast \Pi_{L}\left(jk\right)\, , 
\end{equation}
where $\bm{B}_{\bm{q}}\left(jk\right) = \fourier{\bm{B}_{\bm{q}}(s)}$, and $\Pi_{L}\left(jk\right)$ \textcolor{black}{is outlined in \eqref{eq:sft_length_window}}. The convolution with the \ac{SFT} of the window function is necessary due to the domain adaption (Sec. \ref{sec:spatial_ft}).
\section{Space-Time Fourier Transform} \label{sec:space_time_ft}
    The \ac{STFT} of the strain field $\bm{\xi}(s, t)$ is defined as
\begin{equation} \label{eq:space_time_ft}
    \bm{\Xi}\left(jk, \, j \omega\right) = \int_{0}^{+\infty} \int_{0}^{L} {\bm{\xi}}(s, \, t) \, e^{-j\left(k s + \omega t\right)} \, \textnormal{d} s \, \textnormal{d} t \, ,
\end{equation}
where $\omega = 2 \pi f \in \mathbb{R}$, and $f  \in \mathbb{R}$ is the time-frequency. 

For the discrete case, let us assume to have $M \times N$ samples of the strain field $\bm{\xi}(n \lambda_s, \, m T_s)$, where $m \in [0, \, M - 1]$ and $T_s$ is the sampling period. 
The Discrete \ac{STFT} can be written as
\begin{equation} \label{eq:discrete_stft}
    \bm{\Xi}\left(jk, j \omega\right) = \sum_{m = 0}^{M - 1} \sum_{n = 0}^{N - 1} \bm{\xi}\left(n \lambda_s, m T_s \right) e^{-j \left( k n \lambda_s + \omega m T_s\right)} \, .
\end{equation}
This analysis provides useful information about the dynamic response of the system and highlights certain components in the spatial spectrum that arise during the transient regime.

By examining the curves \textcolor{black}{$\bm{\Xi}(j \bar{k}, \ j \omega)$}, it is possible to observe the time evolution of each spatial harmonic \textcolor{black}{$\bar{k}$}, providing a useful tool to identify resonance or anti-resonance peaks, as well as to assess the relevance of specific spatial harmonics in the time-frequency domain. This information can be leveraged to discard irrelevant harmonics.

Conversely, the curves \textcolor{black}{$\bm{\Xi}(jk, \ j \bar{\omega})$} represent the spatial harmonics excited by an input with a specific time-frequency \textcolor{black}{$\bar{\omega}$}. These curves are useful for understanding the strain profile in the spatial-frequency, offering insights into how the shape of the \ac{CSR} changes in response to an input of a specific time-frequency.

Moreover, if the \ac{CSR} is integrated within a control framework operating at a specific frequency, analyzing \textcolor{black}{$\bm{\Xi}(jk, \, j \bar{\omega})$} allows the user to select an optimal set of basis functions.

\subsection{Dynamics of a Cosserat Rod in the Space-Time Frequency Domain} \label{space_time_ft:dynamics_stft}
From the properties of the 2D \ac{FT} \eqref{eq:fourier_definition}-\eqref{eq:fourier2d_product}, the dynamics of an elastic rod \eqref{eq:crt_dynamics} can be described with a set of algebraic equations in the space-time frequency domain, such as
\begin{equation} \label{eq:dynamics_ft}
    \begin{split}
        &\bm{M} \ast \left(j \omega \, \bm{\mathcal{H}}\right) + \textnormal{ad}_{\bm{\mathcal{H}}}^{*} \ast \left(\bm{M} \ast \bm{\mathcal{H}}\right) = \\
        &jk \left(\bm{F}_i - \bm{F}_a\right) + \textnormal{ad}_{\bm{\Xi}}^{*} \ast \left(\bm{F}_i - \bm{F}_a\right) + \bm{F}_e
    \end{split} \, ,
\end{equation}
where $\bm{M}(jk) = \fourier{\bm{\mathcal{M}}(s)}$, and $\bm{\mathcal{H}}(jk, \, j \omega) = \fourier{\bm{\eta}(s, t)}$.
Moreover, the mixed derivative equality \eqref{eq:mixed_derivative} holds also in the frequency domain, such as 
\begin{equation} \label{eq:mixed_derivative_ft}
    jk \, \bm{\mathcal{H}} = j \omega \, \bm{\Xi} - \left(\textnormal{ad}_{\bm{\Xi}} \ast \bm{\mathcal{H}}\right) \, .
\end{equation}
In \eqref{eq:dynamics_ft}, the \ac{STFT} of the internal wrenches \eqref{eq:internal_passive}, \eqref{eq:internal_active} can be expressed as
\begin{equation} \label{eq:internal_passive_ft}
    \bm{F}_i(j k , \, j \omega) = \bm{\Sigma}(jk) \ast \left(\bm{\Xi} - \bm{\Xi}^{*}\right) + \bm{\Psi}(jk) \ast \left(j \omega \bm{\Xi}\right) \, ,
\end{equation}
\begin{equation} \label{eq:internal_active_ft}
    \bm{F}_a\left(jk, \, j \omega\right) = \bm{B}_{\bm{\tau}}\left(\bm{\Xi}, jk\right) \ast \bm{T}(j \omega) \, ,
\end{equation}
where $\bm{\Psi}(jk) = \fourier{\bm{\Psi}(s)}$, and $\bm{T}(j \omega) = \fourier{\bm{\tau}(t)}$.

Finally, in the case of \ac{GVS} parameterization, we can exploit the separation property of the 2D \ac{FT} \eqref{eq:fourier2d_product} in \eqref{eq:gvs_strain}, resulting in
\begin{equation} \label{eq:gvs_stft}
    \bm{\Xi}(jk, \, j\omega) = \left(\bm{B}_{\bm{q}}\left(jk\right) \, \bm{Q}(j \omega) + \bm{\Xi}^{*}(jk)\right) \ast \Pi_{L}\left(    jk \right) \, ,
\end{equation}
where $\bm{Q}(j \omega) = \fourier{\bm{q}(t)}$. 
With this relation, it is possible to study separately the space and time spectra.
\section{Data-driven Spectrum Extraction} \label{sec:spectrum_extraction}
\begin{figure*}
    \centering
    \includegraphics[width=1.0\linewidth]{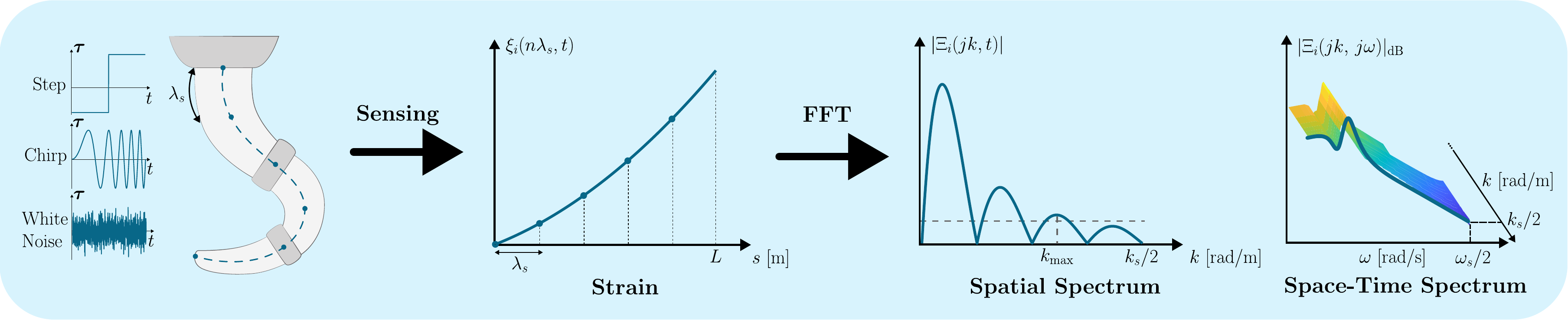}
    \caption{Illustration of the proposed data-driven methodology. The robot is subjected to the standard signals and the samples of the strain field are measured by the sensors. Through \ac{FFT}, the space-time spectrum can be computed.}
    \label{fig:procedure}
\end{figure*}
As outlined in Sec. \ref{sec:spatial_ft} and \ref{sec:space_time_ft}, meaningful insights into the strain field can be derived from both the spatial \eqref{eq:continous_spatial_fourier_transform} and space-time \eqref{eq:space_time_ft} spectra.
This section presents a data-driven approach for extracting the \ac{SFT} and \ac{STFT}, directly from the real-world robots. 
This method only requires knowledge of the robot's length and is independent of its physical parameters.
The proposed procedure consists of the steps exposed below and summarized in Fig. \ref{fig:procedure} \textcolor{black}{and in Alg.~\ref{alg:sofft_procedure}}.
\begin{enumerate}
    \item \textit{Sensorization}: The first step is to sensorize the \ac{CSR}, in order to measure the samples $\bm{\xi}(n \lambda_s)$. The choice of $\lambda_s$ is crucial as it determines the maximum spatial frequency of the sampled spectrum, i.e., $\nu_s / 2$.
    
    \item \textit{Motor Babbling}: Standard signals are applied to the actuators $\bm{\tau}(t)$. Common signals in identification literature include step, chirp, and white noise, thanks to their properties in frequency.
    As a result, samples $\bm{\xi}(n \lambda_s, \, m T_s)$ are obtained, where $T_s$ represents the sampling period of the sensing framework.

    \item \textit{Compute Spectra}: From the samples $\bm{\xi}(n \lambda_s, \, m T_s)$, we compute a time-series of the \ac{SFT} and \ac{STFT} using the \acf{FFT} algorithm. Before this, preprocessing techniques like zero-padding can be applied to improve the resolution of the \ac{FFT}.

    \item \textit{Spectrum Analysis}: From $\bm{\Xi}(jk, \, j \omega)$ we can identify $k_{\textnormal{max}}$ and the shape of the spatial spectrum, which aids in selecting the optimal basis functions. 
    Furthermore, we can derive key dynamic characteristics of the system, such as resonance or anti-resonance phenomena.
    
    \item \textit{Modeling}: 
    Based on the information gathered, the user can determine the minimum number of sections into which the robot can be divided, thereby reducing the need for extensive sensorization ($k_{\textnormal{max}} < \frac{k_s}{2}$). 
    Additionally, optimal basis functions can be extracted using well-established signal processing algorithms such as \ac{MP} \cite{mallat1993matching} or \ac{BPD} \cite{chen2001atomic}.
\end{enumerate}

Viewing strain as a signal offers new insights into the frequency domain of motor babbling \cite{george2020first}. Since standard signals (e.g., white noise) can sample all frequencies in the spectrum, motor babbling can be interpreted as a method for exploring the robot's space-time spectrum. This aligns with the classic definition of motor babbling, where random actuations are applied to the robot to sample its workspace \cite{george2020first}.

\color{black}
\begin{algorithm}
\caption{\textcolor{black}{Data-driven Spectrum Extraction Methodology for \acp{CSR}}}
\label{alg:sofft_procedure}
\begin{algorithmic}[1]
    \color{black}
    \Require Robot length $L$, sampling wavelength $\lambda_s$ and period $T_s$, excitation input $\bm{\tau}(t)$
    \Statex \textbf{Step 1: Sensorization}
    \State Sensorize the \ac{CSR} to measure strain samples $\bm{\xi}(n \lambda_s)$.
    \Statex \textbf{Step 2: Motor Babbling}
    \State Apply standard excitation signals $\bm{\tau}(t)$ to actuators (e.g., step, chirp, white noise).
    \State Record $\bm{\xi}(n \lambda_s, m T_s)$.
    \Statex \textbf{Step 3: Compute Spectra}
    \State $\bm{\Xi}(jk, j\omega) = \text{FFT}\left\{\bm{\xi}(n \lambda_s, m T_s)\right\}$
    \Statex \textbf{Step 4: Spectrum Analysis}
    \State Extract highest spatial frequency $k_{\textnormal{max}}$ and spatial spectrum shape.
    \State Identify key dynamic characteristics (e.g., resonance/anti-resonance).
    \Statex \textbf{Step 5: Modeling}
    \State Determine minimum sections required for the robot using $k_{\textnormal{max}} < \frac{k_s}{2}$.
    \State Extract optimal basis functions via \ac{BPD}.
    \State \Return $k_{\textnormal{max}}$, optimal basis functions, sections, dynamic characteristics.
    \color{black}
\end{algorithmic}
\end{algorithm}
\color{black}

\color{black}
\subsection{Input and External Forces Dependency}
Eq. \eqref{eq:dynamics_ft} shows that the \ac{STFT} of the strain field depends not only on the mechanical properties of the system (i.e., inertia, stiffness, and damping) but also on the input and external forces.
This relationship is well captured in the \ac{ISP} formulation \cite{renda2024dynamics}, where the strain field is expressed as
\begin{equation} \label{eq:implicit_strain}
\bm{\xi}\left(s\right) = \bm{\Sigma}^{-1}\left(\bm{B}_{\bm{\tau}} \bm{\tau} + \bm{B}_{p} \, \bm{q}_p + \bm{B}_{\textnormal{ext}}\right) + \bm{\xi}^{*}(s) \, ,
\end{equation}
where $\bm{B}_{p} \left(\bm{\xi}, \, s\right)\bm{q}_p$ and $\bm{B}_{\textnormal{ext}} \left(\bm{\xi}, \, s\right)$ represent the strain field contributions due to the deformation in the dynamic regime and external forces, respectively. The term $\bm{B}_{\bm{\tau}}\left(\bm{\xi}, \, s\right) \bm{\tau}$ accounts for the deformation excited by actuation in the static regime.

Together, these components determine the deformation modes excited in the system and, consequently, the resulting spectral content. By analyzing the output \ac{STFT}, the strain field can be represented in an optimal basis that captures the combined influence of actuation, external forces, and dynamic effects, extending the static strain analysis  \cite{renda2018discrete}.

Since \acp{CSR} are highly nonlinear systems, the input magnitude can strongly influence the resulting spectrum, introducing nonlinear distortions in the response. These distortions generate additional harmonics that may become particularly evident within specific time–frequency ranges.
For this reason, it is not possible to define a transfer function for such nonlinear systems; instead, their behavior can be described through Volterra series \cite{flake1963volterra, boyd1984analytical}.

In addition, actuator dynamics shape the output spectrum by suppressing harmonics that, while theoretically present, cannot be realized due to physical limitations. As the proposed methodology is applied downstream of the actuation stage, the resulting \ac{STFT} inherently reflects only the harmonics that are physically attainable with the actual actuators.

Conversely, in control applications, the excitation signal may be tailored to probe selected frequency ranges. The corresponding curves $\bm{\Xi}\left(jk, \, j \bar{\omega}\right)$ then provide meaningful information on how the system’s shape varies with the input’s time–frequency content.

\color{black}
\subsection{\ac{BPD} for Strain Fitting} \label{spectrum_extraction:bpd}
The Basis Pursuit Denoising (BPD) problem enables the identification of an optimal basis that preserves the signal's sparsity while mitigating the influence of noise. 
This method is particularly well-suited for applications in soft robotics, where a trade-off between accuracy and reduction in \ac{DoFs} is essential. Specifically, \ac{BPD} selects the best combination of basis functions from a predefined signal dictionary to achieve an optimal reconstruction of the input data.

To adapt the \ac{BPD} problem to the \ac{GVS} approach, the optimization problem can be formulated as
\begin{equation} \label{eq:bpdn_opt_problem}
    \bm{q} = \underset{\bm{q}}{\arg \min} \left\{\frac{1}{2}\norm{\bm{\xi} - \bm{B}_{\bm{q}} \, \bm{q}}^{2}_{2} + \norm{\bm{\gamma} \odot \bm{q}}_1\right\} \, ,
\end{equation}
where $\odot$ is element-wise multiplication, $\norm{\cdot}_i$ is the $\mathcal{\ell}^i$-norm, the basis function matrix $\bm{B}_{\bm{q}}$ represents the signal dictionary and $\bm{\xi}$ the noisy data. Finally, the parameter $\bm{\gamma} \in \mathbb{R}^{n_q}$ controls the trade-off between sparsity and reconstruction accuracy, allowing the user to prioritize either a more compact representation or a more precise fit.

\color{black}
Notably, in \eqref{eq:bpdn_opt_problem}, the strain is assumed to be generated by the basis functions as a linear combination of the coefficients $\bm{q}$. This assumption is well-established in the literature and is shared by similar methods such as \ac{POD}~\cite{alkayas2024soft}.
\color{black}

To evaluate the relevance of each basis, it is possible to adapt the truncation index \eqref{eq:discrete_truncation_criterion} in the continuous case.
\textcolor{black}{Let $b_i(s) \in \mathbb{R}^{n_q}$ the $i$-th column of the basis function matrix.}
The energy associated with $b_i(s)$ can be computed as $E_i(q_i) = q_i^2 \, \frac{1}{2 \pi}\int_{-\infty}^{+\infty} |b_i(jk)|^{2} \, \textnormal{d}k$, where $b_i(jk) = \fourier{b_i(s)}$.
Therefore, the energy fraction relative to the total reconstructed strain field energy is
\begin{equation} \label{eq:truncation_index}
    E_{\textnormal{tr}, b_i}(q_i) = \frac{E_i(q_i)}{\sum\limits_{j = 0}^{n - 1} E_j(q_j)} \, .
\end{equation}

It is worth highlighting that, thanks to the continuous \ac{SFT}, the energy evaluation can be performed for all the wavenumbers.
\color{black}
However, the energy can still be computed numerically using the discrete Parseval identity \eqref{eq:discrete_parseval}, for any chosen range and resolution of wavenumbers.

Finally, the user can perform the truncation by discarding the basis functions whose truncation indices are below a specified threshold.
\color{black}

\section{Numerical Validation} \label{sec:numerical_validation}
\begin{figure}
    \centering
    \includegraphics[width=1.0\linewidth]{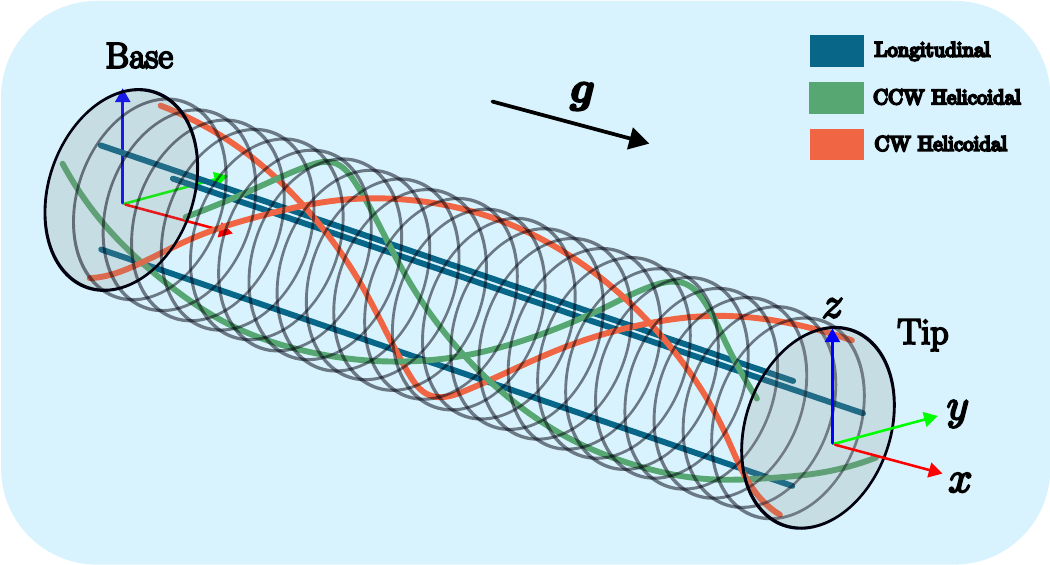}
    \caption{Sketch of the H-Support robot, a cylindrical \ac{CSR} with 3 longitudinal and 4 helicoidal actuators. The geometrical and physical parameters are listed in Tab. \ref{tab:sim_parameters}.}
    \label{fig:h_support}
\end{figure}

    The data-driven procedure is validated through numerical examples. We applied this method to two simulated robots.
The former is H-Support (Fig. \ref{fig:h_support}), a \ac{CSR} with a cylindrical cross-section, actuated by three longitudinal and four helicoidal actuators. This robot is a modified version of the existing I-Support \cite{manti2016soft}.
Table \ref{tab:sim_parameters} lists the geometrical and physical parameters.
With the same physical characteristics, the latter robot is the conical variant of the H-Support (Fig. \ref{fig:conical_hsupport_sketch}).
\begin{table}
\centering
\caption{Geometrical and physical parameters of the Simulated H-Support}
\label{tab:sim_parameters}
    \begin{tabular}{lll}
    \toprule
    Name                    &   Symbol          &   Value                                                               \\
    \midrule
    Length                  &   $L$               &   \SI{1.0}{\meter}                                                            \\
    Cross-Section Radius               &   $\bar{R}_{\textnormal{cs}}$               &   \SI{0.1}{\meter}                                                            \\
    Density                 &   $\bar{\rho}$    &   \SI{1000.0}{\kilogram/\meter^3}                                     \\     
    Young's Modulus         &   $E$             &   \SI{1.0}{M\pascal}                                             \\
    Poisson Ratio           &   $\nu$           &   0.5                                                                     \\
    Damping Coefficient                &   $\beta$         &   \SI{0.01}{M\pascal \cdot \second}         \\
    Stress-Free Strain      & $\bm{\xi}^{*}$    &   $[0, \, 0, \, 0, \, 1, \, 0, \, 0]^{\top}$    \\
    \bottomrule
    \end{tabular}
\end{table}

\color{black}
The simulations were conducted using the MATLAB toolbox \ac{SoRoSim} \cite{mathew2022sorosim}, which implements \ac{GVS} and is therefore based on \ac{CRT}. This simulator has been validated in \cite{mathew2022sorosim} against high-accuracy \ac{FEM}-based softwares, showing an excellent trade off between accuracy and computational efficiency. 
Since \ac{SoRoSim} relies on \ac{GVS}, the strain basis remains fixed, a feature that is expected to be reflected in the resulting output \ac{STFT}.
\color{black}

The robots are simulated with three different functional bases: (i) polynomial \eqref{eq:polynomial_basis}, (ii) trigonometrical \eqref{eq:fourier_basis}, and (iii) Gaussian \eqref{eq:gaussian_basis}. 
In all cases, we employed a second-order truncation, \textcolor{black}{i.e., we truncated the summations in \eqref{eq:polynomial_basis} - \eqref{eq:gaussian_basis} after the second term.}
For each basis, we followed the procedure outlined in Sec. \ref{sec:spectrum_extraction}, \textcolor{black}{to compute the \ac{STFT} of the resulting strain field}.

The strain field was sampled with a sampling wavenumber of $\nu_s = \SI{100}{\meter^{-1}}$, corresponding \textcolor{black}{to} a sampling wavelength of $\lambda_s = \SI{0.01}{\meter}$.
\color{black}
The \acp{CSR} were tested using the following input:
\begin{equation} \label{eq:chirp_simulations}
\bm{\tau}\left(t\right) = \left[K \sin\left(2 \pi t^2 + i \frac{2 \pi}{n_a}\right) \right]_{i = 0}^{n_a - 1} \, .
\end{equation}
For the H-Support, the gain was set to $K = 10^{3}$, while for the Conical H-Support it was reduced to $K = 2.5 \cdot 10^{2}$, since its geometry produces sufficiently large deformations that can lead to nonlinear distortions. The robot was simulated for $\SI{100}{\second}$, with the chirp signal reaching a maximum frequency of $\SI{100}{\hertz}$. Simulations were performed using the ode15s solver, an advanced implicit method suitable for stiff \acp{ODE}, and the results were resampled at $f_s = \SI{1}{k \hertz}$, yielding the samples $\bm{\xi}(n \lambda_s\, , m T_s)$.

To estimate the nonlinear distortion, we first linearized the system around the equilibrium point for $\bm{\tau} = \bm{0}$. The matrices of the linearized system \eqref{eq:linearized_system}-\eqref{eq:B_lin} were obtained using the analytical derivatives recently implemented in the differentiable version of \ac{SoRoSim} \cite{mathew2024analytical}.
Once the linearized system was computed, we applied \eqref{eq:chirp_simulations} to evaluate the difference between the original and linearized systems, thus estimating the nonlinear distortions.
\color{black}

\subsection{H-Support} \label{numerical_validation:hsupport}
\begin{figure*}
        \centering
        \includegraphics[width=1.0\linewidth]{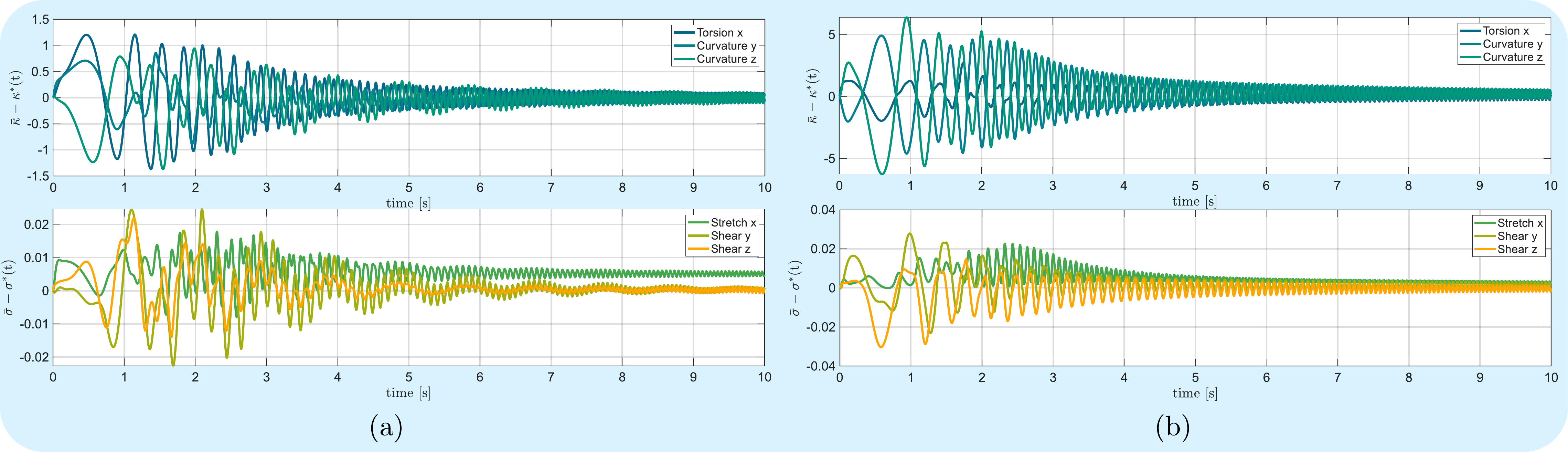}
        \caption{\textcolor{black}{Time evolution of the spatially averaged deformation. Simulations of the H-Support (a) and Conical H-Support (b) are presented using the polynomial basis.}}
        \label{fig:avg_xi}
    \end{figure*}

        \color{black}
    Fig. \ref{fig:avg_xi}a shows the time evolution of the spatial average strain\footnote{\textcolor{black}{The spatial average strain is computed by $\bar{\bm{\xi}} - \bm{\xi}^{*}$, where $\bar{\bm{\xi}} = \frac{1}{N} \sum^{N - 1}_{n = 0} \bm{\xi}\left(n \lambda_s\right)$.}}, illustrating the range of deformations for both angular and strain modes. It is noteworthy that the average over space corresponds to the harmonic at $\nu = \SI{0}{\meter^{-1}}$.
    
\begin{figure}
        \centering
        \includegraphics[width=1.0\linewidth]{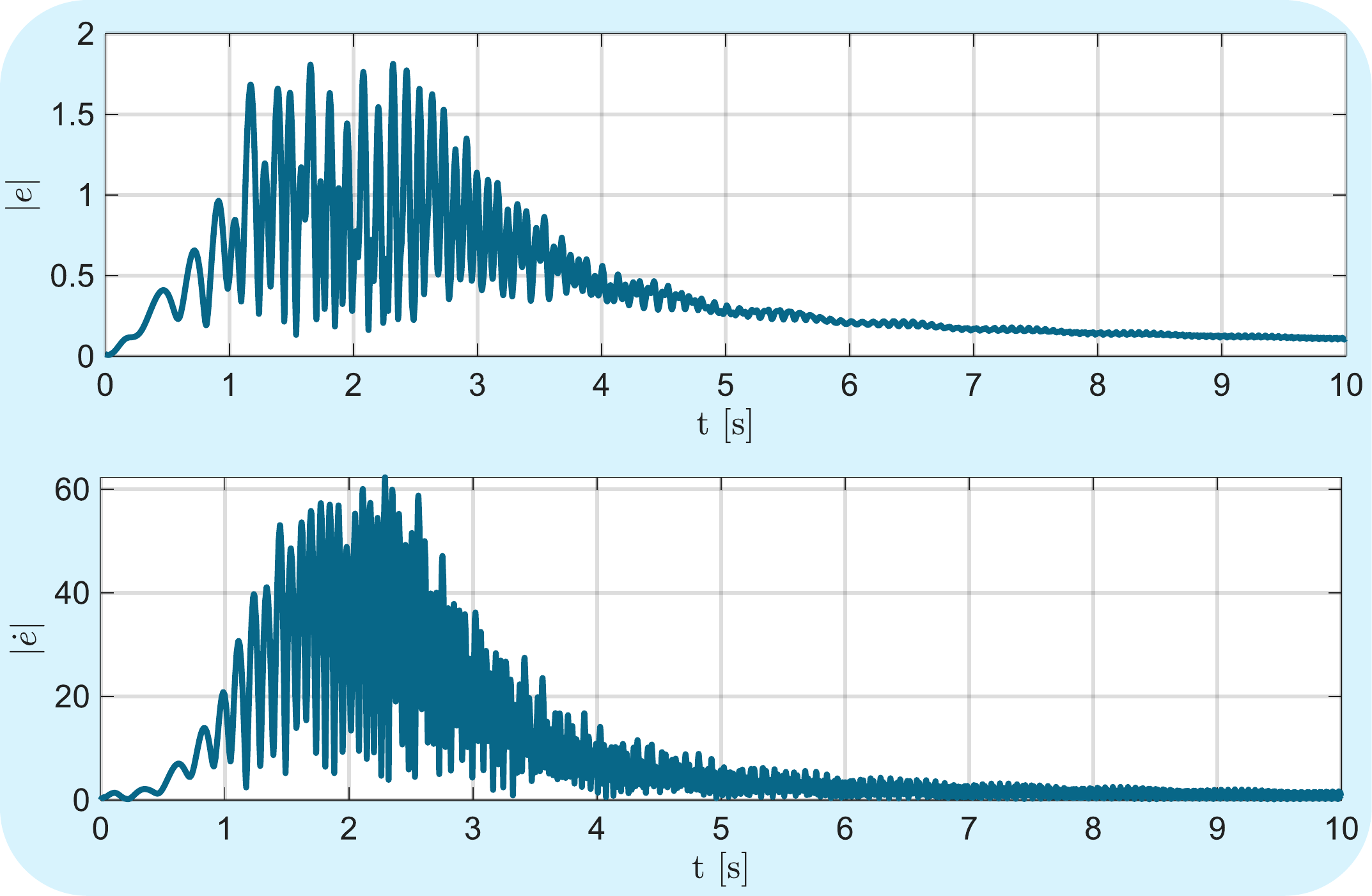}
        \caption{\textcolor{black}{Error between the nonlinear and linearized systems, computed as the norm of the differences in $\bm{q}$ and $\dot{\bm{q}}$.}}
        \label{fig:nonlinearity_time}
    \end{figure}

    In addition, Fig. \ref{fig:nonlinearity_time} shows the discrepancy between the nonlinear and linearized systems in terms of the norm of the error in $\bm{q}$ and $\dot{\bm{q}}$. The discrepancy remains significant up to approximately $\SI{5}{\second}$, after which both the nonlinear and linearized systems start to neglect high-frequency oscillations. Between $\SI{1}{\second}$ and $\SI{3}{\second}$, the maximum discrepancy occurs, corresponding to the system’s main resonance. Since the linearization is effective only for small $\dot{\bm{q}}$, the resonance induces an increase in $\dot{\bm{q}}$, making the Coriolis effect more prominent and generating the nonlinear distortions shown in Fig. \ref{fig:nonlinearity_time}.
    \color{black}
    

\begin{figure*}
        \centering
        \includegraphics[width=1.0\linewidth]{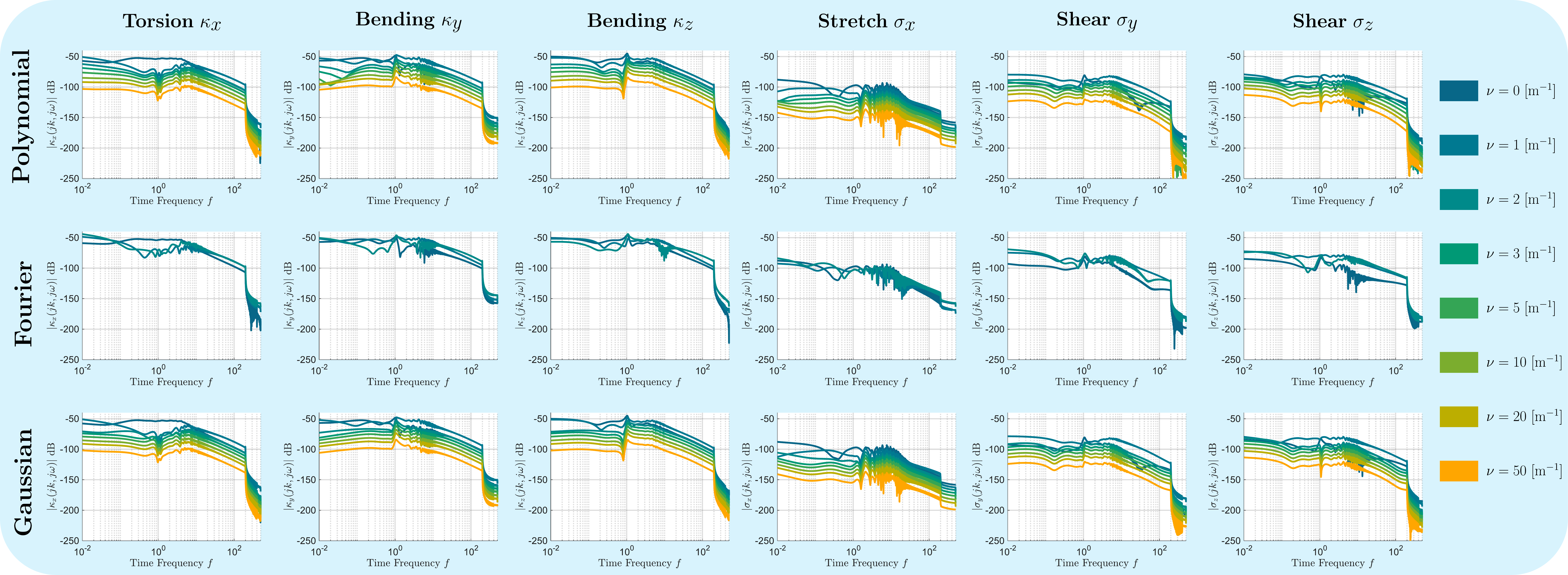}
        \caption{\textcolor{black}{The space-time spectra of the H-Support numerical example discussed in Sec. \ref{numerical_validation:hsupport}. The \acp{STFT} are organized in functional basis (rows) and strain modes (columns).}}
        \label{fig:stft_time}
    \end{figure*}

    In Fig. \ref{fig:stft_time} is shown the \acp{STFT} of the three simulations, organized in functional basis (row) and strain modes (columns). The results are discussed in the following subsections.
    \subsubsection{Spatial Spectrum Components}
    \begin{figure}
        \centering
        \includegraphics[width=1.0\linewidth]{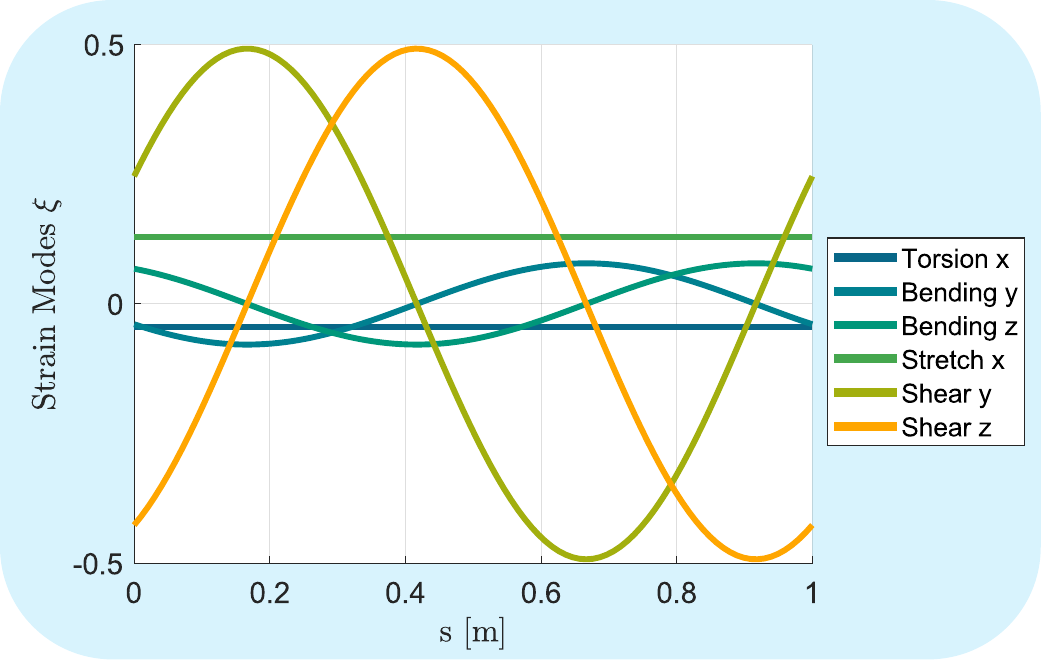}
        \caption{Strain Analysis of the H-Support robot with only an active helicoidal actuator with a magnitude of $\SI{1}{\newton}$. For bending and shear, the numerical analysis reveals a trigonometric pattern; for twisting and shear, a constant pattern.}
        \label{fig:strain_analysis}
    \end{figure}
    From Fig. \ref{fig:stft_time}, we can infer the spatial spectrum composition for each strain mode.

    Regarding the twisting mode ($\kappa_x$), the most prominent component is the constant term across each functional basis.
    In contrast, the bending and shear modes exhibit distinct spectra. Counterintuitively, the harmonic at $\nu = 0 \; \textnormal{m}^{-1}$ is significantly less prominent than the subsequent harmonics. This behavior can be attributed to the helicoidal actuators, which excite the bending and shear modes with a sinusoidal profile along the rod.
    
    To validate this observation, we performed \textcolor{black}{static} strain analysis using the \ac{ISP} method \cite{renda2020geometric, renda2024dynamics}. Specifically, the strain field can be numerically computed by solving
    \begin{equation} \label{eq:static_implicit_strain}
        \bm{\xi}(s) = \bm{\Sigma}^{-1}(s) \, \bm{B}_{\bm{\tau}}\left(\bm{\xi}, s\right) \bm{\tau} + \bm{\xi}^{*}(s) \, .
    \end{equation}
    
    The solution of \eqref{eq:static_implicit_strain} corresponds to the strain modes excited by the actuators in the static regime \cite{renda2020geometric}.
    Fig. \ref{fig:strain_analysis} shows the results \textcolor{black}{for an helicoidal actuator}, highlighting a constant stretch and twist. 
    Moreover, the analysis reveals that the bending and shear strains exhibit a sinusoidal profile. Notably, the bending and shear strains concerning the same axis (i.e., $\kappa_y, \sigma_y$ and $\kappa_z, \sigma_z$) are in phase opposition. 
    \color{black}
    This phase opposition arises naturally from the rod’s static equilibrium under the helicoidal tendon’s pull: in regions where bending is minimal, the shear strain increases to balance the internal forces, resulting in the observed pattern.
    \color{black}
    
    Furthermore, this finding reveals a coupling between the bending and shear modes in static conditions, which arises from the specific geometry of the actuator.

    \subsubsection{Resonance and Anti-resonance Peaks}
    \color{black}
    Regardless of the selected functional basis, the system exhibits a sequence of resonance and anti-resonance peaks over time, a behavior characteristic of systems with a high number of degrees of freedom (DoFs) \cite{Ewins1999}.
    
    These resonance and anti-resonance peaks are intrinsically associated with vibrational phenomena, defining the conditions under which the system undergoes constructive or destructive interference in response to a sinusoidal input. In the case of the curves $\bm{\Xi}\left(j \bar{k}, \, j\omega\right)$, each spatial harmonic $\bar{\nu}$ possesses a distinct dynamic behavior, characterized by specific resonance and anti-resonance frequencies.

    The differences among spatial harmonics arise from the choice of strain basis, which governs the distribution of deformation energy across spatial modes. 
    For instance, while the polynomial basis distributes energy according to an inverse proportional law ($\propto \frac{1}{jk^{h + 1}}$, where $h$ denotes the order), the Fourier basis concentrates deformation energy in selected harmonics—specifically, in this case, $\nu = \SI{1}{\meter^{-1}}$ and $\SI{2}{\meter^{-1}}$.

    The system exhibits a principal resonance peak at approximately $\SI{1}{\hertz}$, particularly evident in the bending modes. Unlike in the static regime, the constant component of the bending mode becomes significant at the harmonic $\nu = \SI{1}{\meter^{-1}}$. All harmonics display comparable resonance peaks, resulting in complex deformation patterns during these vibrations.
    
    At the same frequency, the higher-order harmonics of $\kappa_x$ experience a pronounced reduction in magnitude, whereas the constant component (i.e., $\nu = \SI{0}{\meter^{-1}}$) reaches its maximum amplitude. This observation indicates that the torsion remains predominantly constant along the rod during vibration of the resonance.
    
    A similar trend is observed in the shear strain, which exhibits a main resonance around $\SI{1}{\hertz}$. In this frequency range, the constant shear component becomes prevalent, in contrast to its behavior at other frequencies. 
    This phenomenon can be attributed to the dominance of longitudinal actuation during resonance, as similarly observed in the bending response.
    
    As previously discussed, the selection of the strain basis significantly influences the temporal–spatial dynamics of the harmonics. This effect is most evident in the higher-order harmonics, while the constant component exhibits comparable dynamics across different bases. 
    The Gaussian basis produces a spectrum closely resembling that of the polynomial basis, whereas the Fourier basis yields a substantially different distribution, concentrating nearly all deformation energy within the first two harmonics.
    
    Although the principal resonance remains located at $\SI{1}{\hertz}$, the spectrum reveals multiple secondary resonance and anti-resonance peaks, indicating a richer vibrational behavior compared to the other bases. This suggests the presence of coupled vibrations, where spatial deformation propagating along the rod interacts with the temporal oscillations of the system.
    
    \color{black}
    
    \subsubsection{Frequency-varying Behavior}
\begin{figure}
        \centering
        \includegraphics[width=1.0\linewidth]{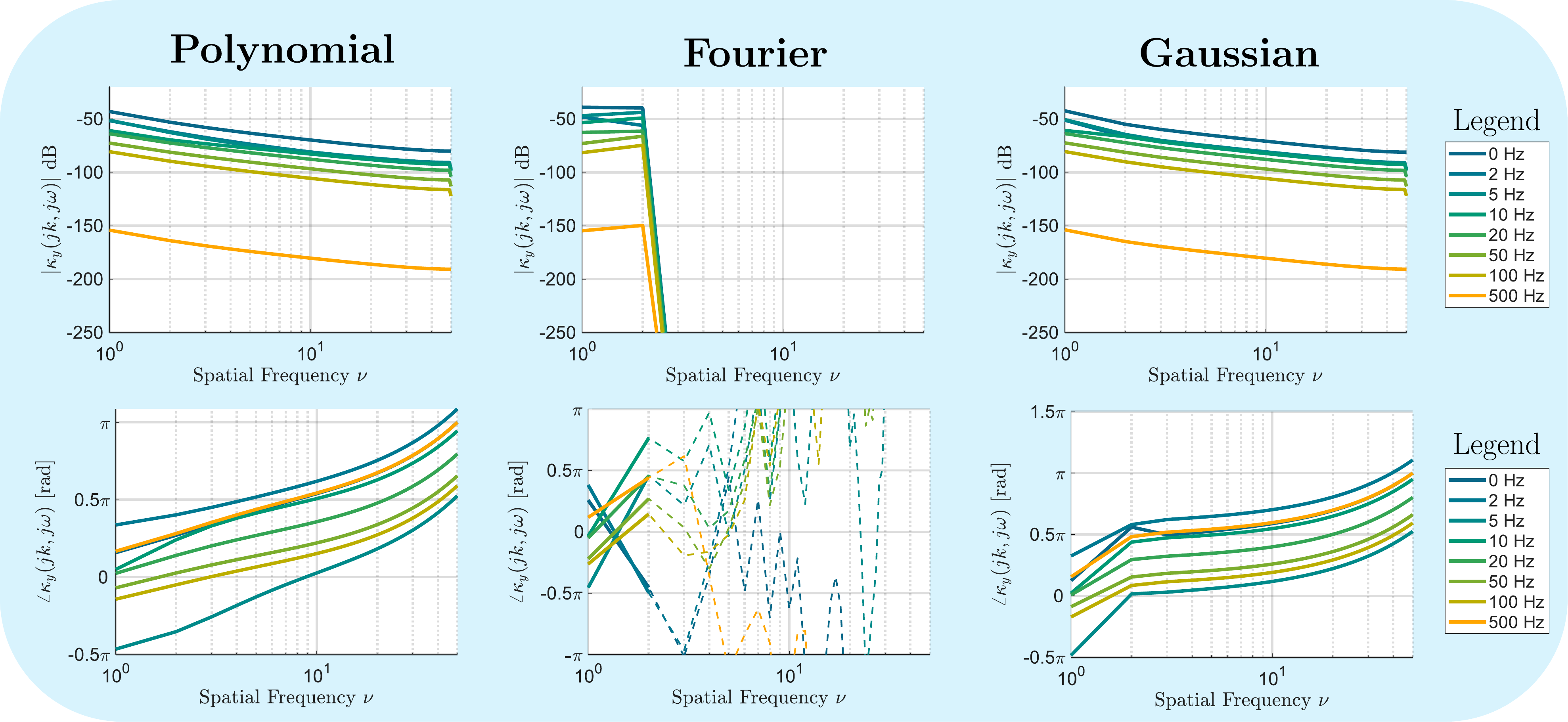}
        \caption{\textcolor{black}{Spatial Spectrum varying the time-frequencies.    
        The spatial harmonics $\nu > \SI{2}{\meter^{-1}}$ are not taken into account in the case of trigonometric basis since the analysis is limited to the second order.}}
        \label{fig:stft_space}
    \end{figure}

    As discussed in Sec. \ref{sec:space_time_ft}, the \ac{STFT} highlights a well-known characteristic of \acp{CSR}: their deformations vary according to the time–frequency content of the input signal. To illustrate this behavior, Fig. \ref{fig:stft_space} reports the evolution of the spatial spectrum as a function of time–frequency.

    Since the strain field is expressed through a specific functional basis, the shape of the spatial spectrum remains invariant with respect to time–frequency variations; only its magnitude changes accordingly.

    \color{black}
    Furthermore, as previously observed, Fig. \ref{fig:stft_space} clearly illustrates how the selected strain basis governs the distribution of deformation energy across spatial frequencies. In addition, the phase plot shows how each basis distributes the deformation energy along the rod.
    \color{black}

    \color{black}
    \subsubsection{Input Dependency: Nonlinearities and Actuators' Dynamics}
\begin{figure}
        \centering
        \includegraphics[width=1.0\linewidth]{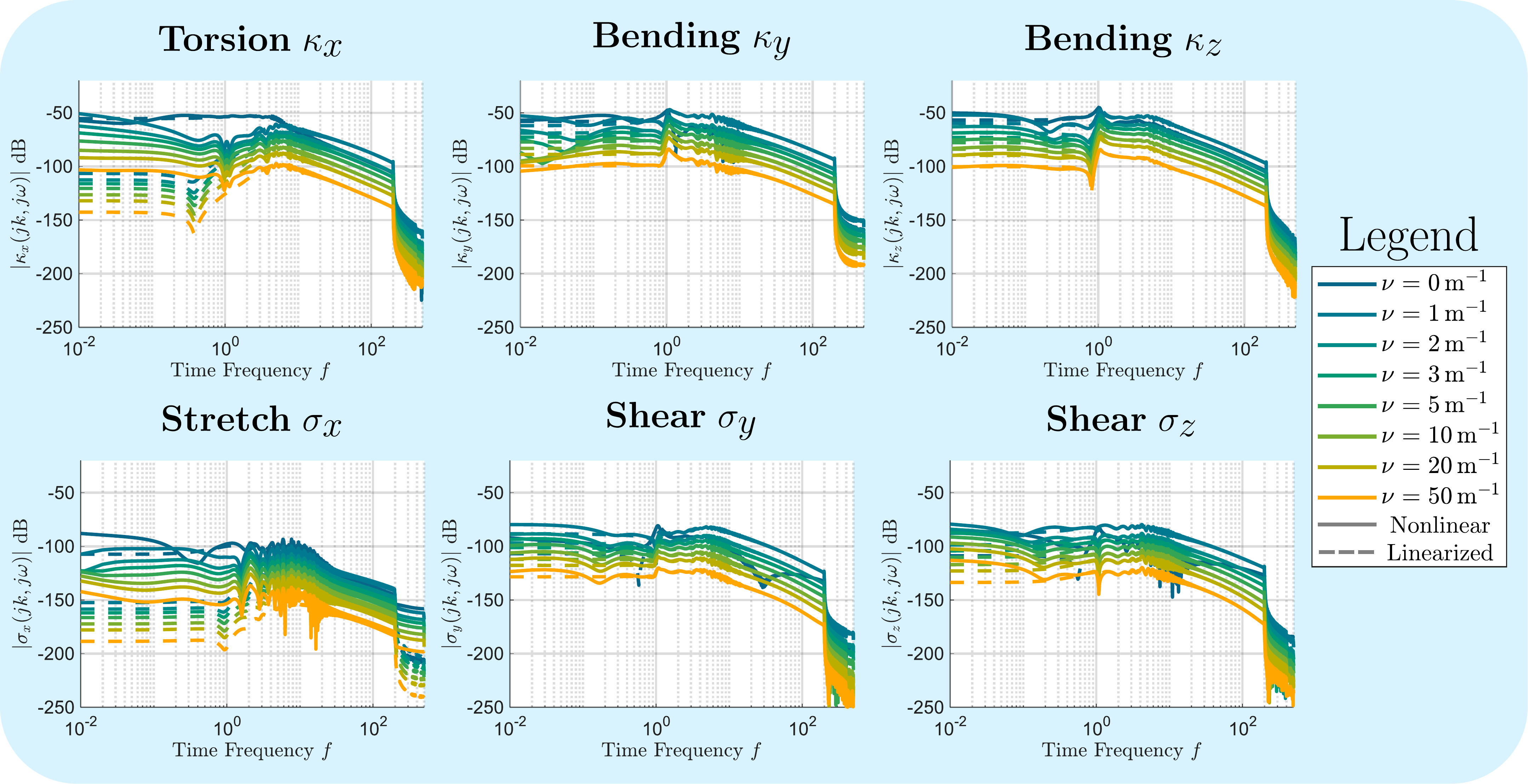}
        \caption{\textcolor{black}{Difference between the nonlinear and the linearized system in the frequency domain.}}
        \label{fig:nonlinearity}
    \end{figure}

    As discussed in Sec. \ref{numerical_validation:hsupport}, the input magnitude is sufficient to excite nonlinear effects in the strain field’s time evolution. Figure \ref{fig:nonlinearity} shows the \ac{STFT} comparison between the original (solid line) and linearized (dotted line) systems presented in Fig. \ref{fig:nonlinearity_time}. In the frequency domain, the most pronounced discrepancies appear below $\SI{3}{\hertz}$, consistent with the observations in the time domain. Beyond this range, the two systems exhibit similar behavior, as the mechanical dynamics naturally attenuate higher-frequency oscillations, thereby reducing the difference between the nonlinear and linearized responses.
    
    Notably, within the frequency band where the discrepancy becomes significant, the deviation affects not only the magnitude but also the overall shape of the spectrum. 
    This occurs because the linearized model fails to capture the nonlinear interactions between bending and shear and the axial deformations (torsion and stretch). Such coupling effects become particularly pronounced for relatively high values of $\norm{\dot{\bm{q}}}$, due to the Coriolis effect.

    Actuator dynamics play an important role in this context, as they effectively act as low-pass filters that suppress high-frequency components. We regard this as a strength of our data-driven approach: it inherently accounts for the spatial and temporal frequencies that remain after the actuator dynamics. Consequently, it avoids modeling spatial frequencies that are theoretically possible but physically unattainable by the real system.
    
\begin{figure}
        \centering
        \includegraphics[width=1.0\linewidth]{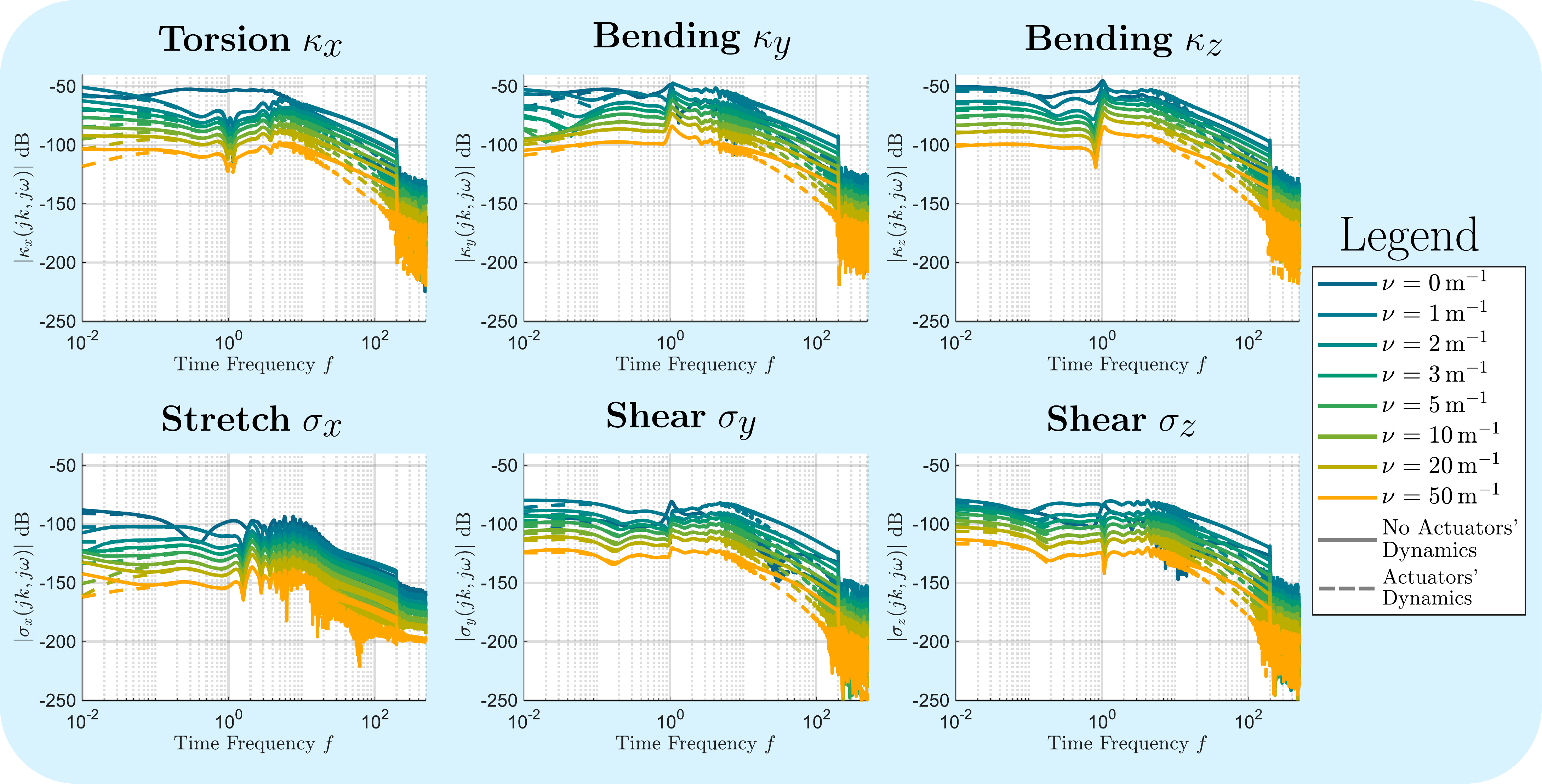}
        \caption{\textcolor{black}{Effect of actuators' dynamics on the output \ac{STFT}, which act as a low-pass filter with a cut-off frequency of $\SI{10}{\hertz}$.}}
        \label{fig:stft_lowpass}
    \end{figure}

    To further investigate the influence of actuator dynamics, we performed an additional simulation in which a transfer function was inserted between the signal generator and the robot to model typical actuator behavior (e.g., DC motors). Specifically,
    \begin{equation}
    \bm{G}_{\textnormal{act}}(j \omega) = \frac{20 \pi}{j \omega + 20 \pi} \bm{I}_{7} \, ,
    \end{equation}
    where the cut-off frequency is $f_c = 10$ Hz and the gain is unity. Figure \ref{fig:stft_lowpass} shows the STFT comparison with and without actuator dynamics, highlighting their filtering effect on the system response.
    \color{black}

\subsection{Conical H-Support} \label{numerical_validation:conical_hsupport}
\begin{figure}
    \centering
    \includegraphics[width=1.0\linewidth]{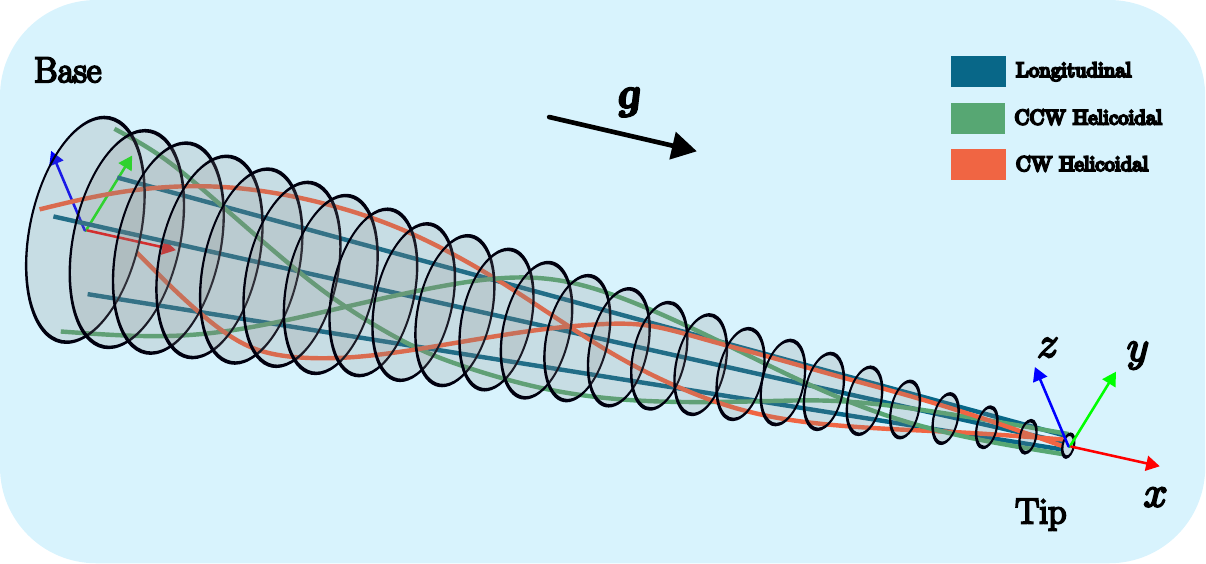}
    \caption{A sketch of the Conical H-Support. The conical shape is described by a $s$-varying cross-section's radius of $R_{\textnormal{cs}}(s) = \bar{R}_{\textnormal{cs}} \left(1 -0.9 s\right)$. The geometrical and physical parameters are listed in Tab. \ref{tab:sim_parameters}.}
    \label{fig:conical_hsupport_sketch}
\end{figure}

The numerical validation is also performed on a conical H-Support variant (Fig. \ref{fig:conical_hsupport_sketch}), where the physical and geometrical parameters are kept the same and listed in Table \ref{tab:sim_parameters}.
The conical profile of the \ac{CSR} is realized by the linear function of the cross-section's radius, such as $R_{\textnormal{cs}}(s) = \bar{R}_{\textnormal{cs}} \left(1 -0.9 s\right)$.
As a consequence, the cross-section's area and second moment of area will be $A(s) \propto s^2$, and $J_i(s) \propto s^4$, respectively.
Therefore, in contrast to the cylindrical case, the dynamic matrices (i.e., inertia, stiffness, and damping) are $s$-dependent, playing an active role on the \ac{STFT} spectra.

\color{black}
In Fig. \ref{fig:avg_xi}b, the average deformation over time of the Conical H-Support under the input \eqref{eq:chirp_simulations} is shown.
Despite the lower gain, the Conical H-Support exhibits greater bending than the standard H-Support. 
This behavior results from the system geometry, reducing the total mass and lowers the stiffness at the tip.
\color{black}

\begin{figure}
    \centering
    \includegraphics[width=1.0\linewidth]{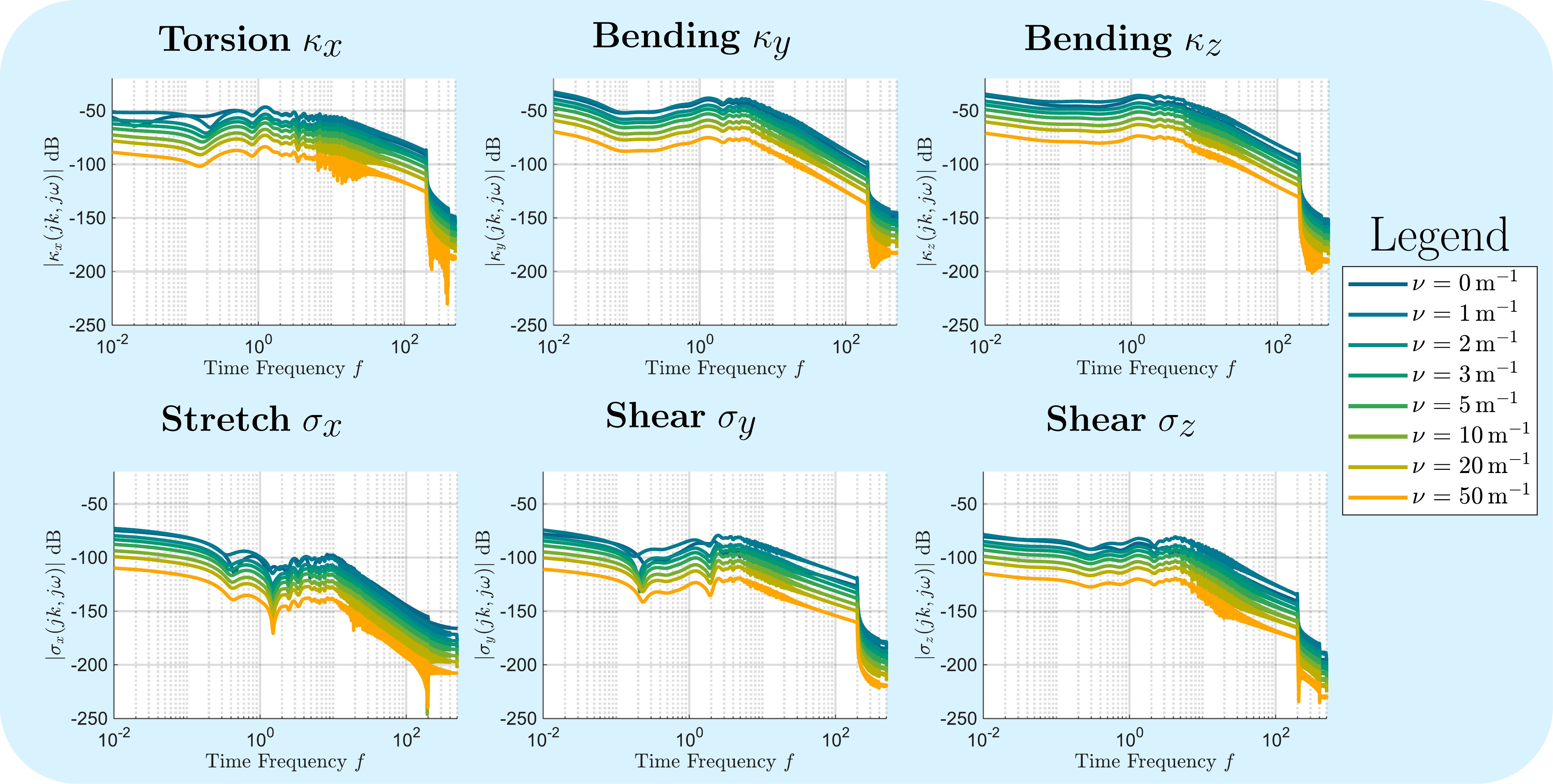}
    \caption{\textcolor{black}{The space-time spectra of the Conical H-Support numerical example discussed in Sec. \ref{numerical_validation:conical_hsupport}.}}
    \label{fig:conical_hsupport_stft}
\end{figure}

Fig. \ref{fig:conical_hsupport_stft} shows the \ac{STFT} for the polynomial basis case.

\color{black}
Differently from the cylindrical case, the absence of clear separation between the constant and higher spatial harmonics of the spectrum indicates that the geometry of the robot promotes a more uniform distribution of the deformation energy. As a result, all harmonics share a similar spectral shape, differing mainly in amplitude, particularly evident in the bending and torsion modes.
\color{black}

\color{black}
\subsection{Strain Fitting with \ac{BPD}}
\begin{figure*}
    \centering
    \includegraphics[width=1.0\linewidth]{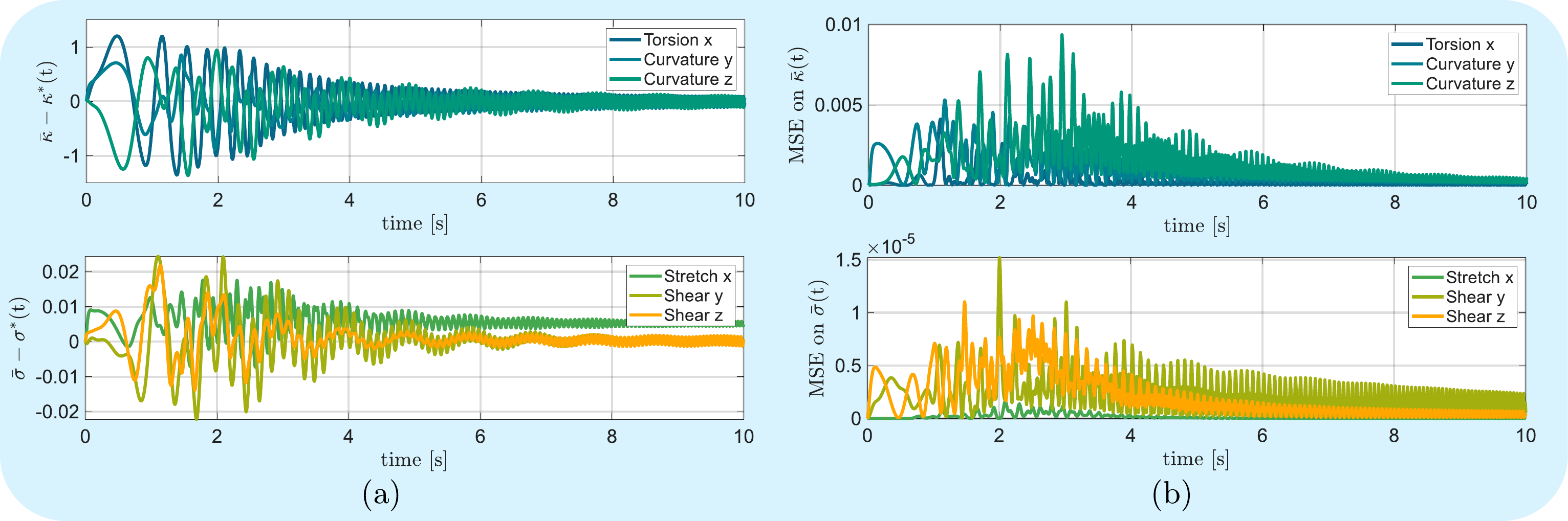}
    \caption{\textcolor{black}{Simulation of the H-Support using a fourth-order polynomial basis. (a) shows the average deformation over time. After computing the optimal truncation using \ac{BPD}, the H-Support is re-simulated with a new mixed basis combining lower-order polynomial and trigonometric functions. (b) shows the \ac{MSE} over time between the two simulations.}}
    \label{fig:fitting_mse}
\end{figure*}

To validate the \ac{BPD} algorithm for strain fitting, we simulated the H-Support modeled using a fourth order polynomial basis with a 30-\ac{DoFs} representation. The time-evolution of the average deformation is shown in Fig. \ref{fig:fitting_mse}a.
A signal dictionary was constructed using polynomial and trigonometric bases, both truncated at first order, to ensure that the algorithm could not reproduce the exact strain basis. 
The \ac{BPD} algorithm was then applied with a sparsity gain of $\bm{\gamma} = [0.5, \, 0.5, \, 0.5, \, 0.07, \, 0.05, \, 0.05]^{\top}$.

\begin{figure}
    \centering
    \includegraphics[width=1.0\linewidth]{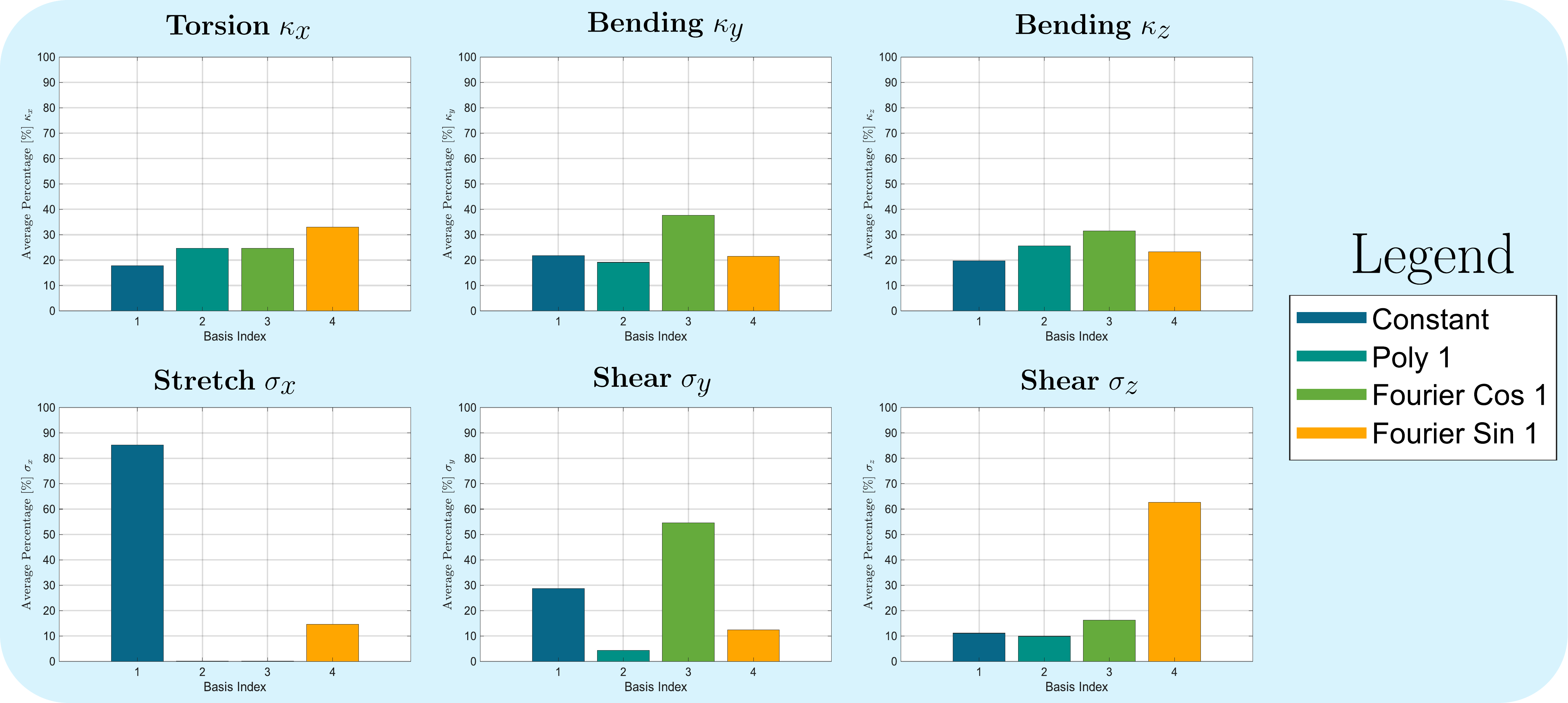}
    \caption{\textcolor{black}{Average truncation index of the bases selected in the signal dictionary.}}
    \label{fig:fitting_4th}
\end{figure}

The average truncation index of the selected bases is shown in Fig.~\ref{fig:fitting_4th}. Despite the lack of higher-order polynomial terms, the trigonometric basis effectively compensates for this limitation. This behavior can be interpreted by noting that trigonometric functions inherently contain higher-order terms when expanded as a Taylor series, e.g., $\sin(s) \approx s - \frac{1}{3!}s^{3} + \frac{1}{5!}s^{5} + \dots$ and $\cos(s) \approx 1 - \frac{1}{2!}s^{2} + \frac{1}{4!} s^{4} + \dots$. Thus, the trigonometric basis implicitly captures higher-order polynomial contributions.
To assess the reconstruction accuracy, we re-simulated the H-Support using the optimal basis obtained from the \ac{BPD}, which resulted in a reduced model with 24~\ac{DoFs}. The system was subjected to the same input excitation \eqref{eq:chirp_simulations}. After computing the strain field, we evaluated the \ac{MSE} between the original and reconstructed strain fields over space. The temporal evolution of the \ac{MSE} is shown in Fig.~\ref{fig:fitting_mse}b.
\color{black}
\section{Experimental Validation} \label{sec:experimental_validation}
\begin{figure}
    \centering
    \includegraphics[width=1.0\linewidth]{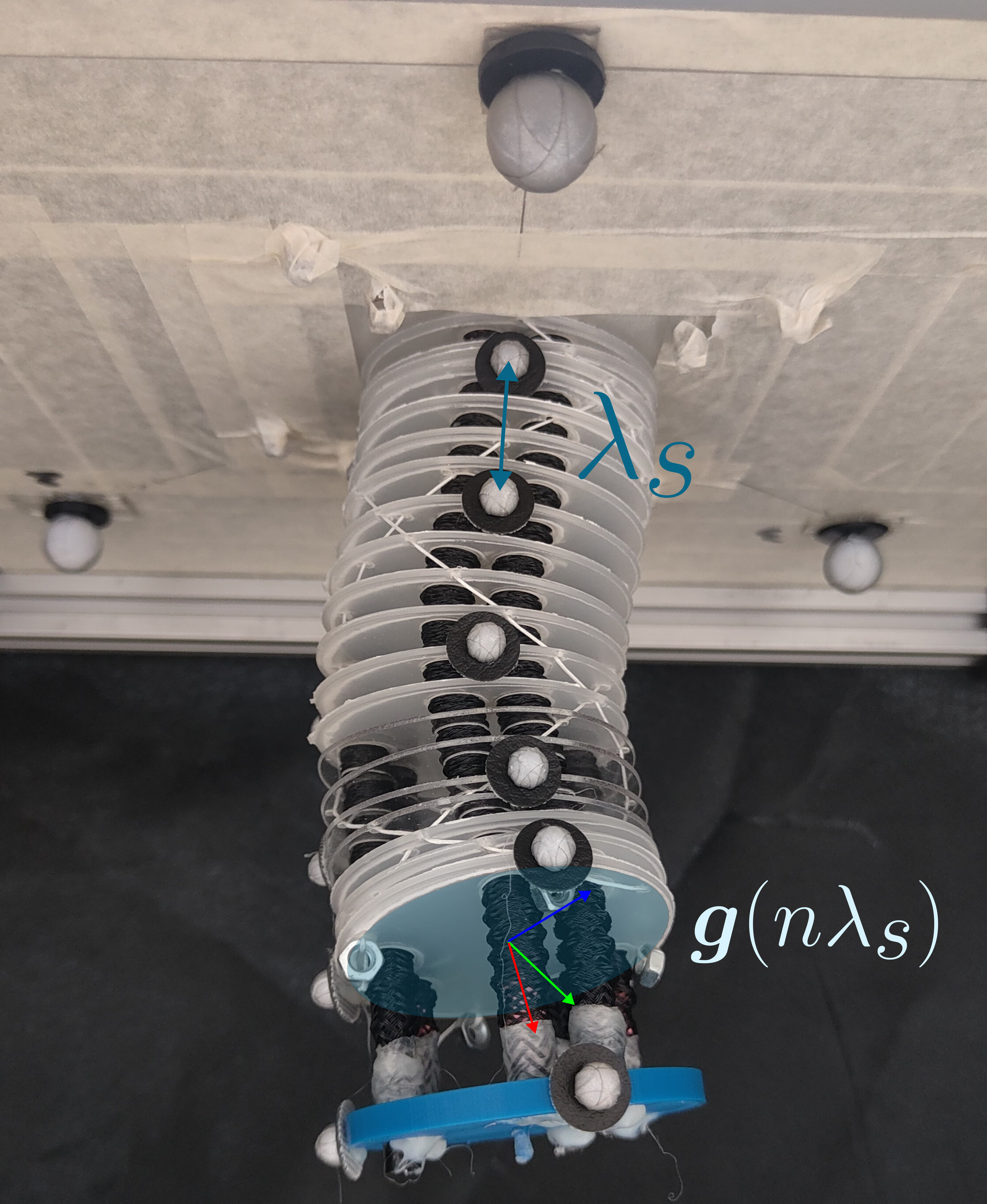}
    \caption{The H-Support prototype is a cylindrical robot with 3 longitudinal pneumatic chambers and 4 tendons arranged with a helicoidal path. The VICON system sensorizes the \ac{CSR}, measuring the positions of the cross-sections each $\lambda_s = \SI{0.032}{\meter}$.}
    \label{fig:h_support_prototype}
\end{figure}

To demonstrate the consistency of the proposed methodology, we conducted an experimental validation using the H-Support prototype (Fig. \ref{fig:h_support_prototype}).
This robot is actuated by three longitudinal pneumatic-driven actuators and four helicoidal tendon-driven actuators.
\color{black}
The longitudinal actuators, described in \cite{manti2016soft}, are capable of operating at pressures up to $1 \; \textnormal{bar}$. The tendons are driven by Dynamixel XM430-W350-R servo motors, which provide a maximum torque of $4.1 \; \textnormal{N} \cdot \textnormal{m}$. 

The actuation sources of the prototype impose inherent physical constraints. Specifically, $\bm{\tau}_{1:3} \geq \bm{0}_3$ indicates that the pneumatic chambers are capable of generating forces that induce extension, whereas $\bm{\tau}_{4:7} \leq \bm{0}_4$ reflects that the tendons can exert forces that produce compression.
To tackle these limitations, the absolute value was applied to \eqref{eq:chirp_simulations}, thereby ensuring that each actuator receives an input with the appropriate sign.
This modification enables the maximization of the deformation achievable within the constraints of the chosen actuation scheme.
The gains are configured to utilize the full actuation range allowed by the current experimental setup.
\color{black}

Concerning sensing, the H-Support is equipped with a Motion Capture system (VICON), which measures the roto-translation of sensorized cross-sections. 
Markers are placed at equal intervals of $\lambda_s = \SI{0.032}{\meter}$ along the robot's length, providing a sampling wavenumber of $\nu_s = \SI{31.25}{\meter^{-1}}$. 

Based on these measurements, the strain field samples can be calculated as
\begin{equation} \label{eq:vicon2strain}
    \bm{\xi}(n \lambda_s) = \frac{1}{\lambda_s}
    \left(\log_{SE(3)}\left(\bm{g}^{-1}\left(n \lambda_s\right) \bm{g}\left((n + 1) \lambda_s\right)\right)\right)^{\vee} ,
\end{equation}
where $\log_{SE(3)}(\cdot)$ represents the logarithmic map in the $SE(3)$ group \eqref{eq:logSE3_definition}. 
The relation \eqref{eq:vicon2strain} is derived by inverting the forward kinematics of a \ac{GVS} model \cite{mathew2024reduced}, considering a second-order Zanna collocation \cite{zanna1999collocation}.
\color{black}
\eqref{eq:vicon2strain} is equivalent to reconstruct the strain field using \ac{ZOH} with a sampling wavelength $\lambda_s$. 
This approach is chosen as it directly computes the strain from the raw data without mainly altering the spectrum. Alternative methods, such as fitting, could negatively affect the calculation of optimal basis functions, since the fitting process essentially pre-imposes a solution structure.
\color{black}

Concerning time samples, the VICON system provides the measures at $f_s = \SI{100}{\hertz}$.

\subsection{Spectrum Analysis}
\begin{figure*}
    \centering
    \includegraphics[width=1.0\linewidth]{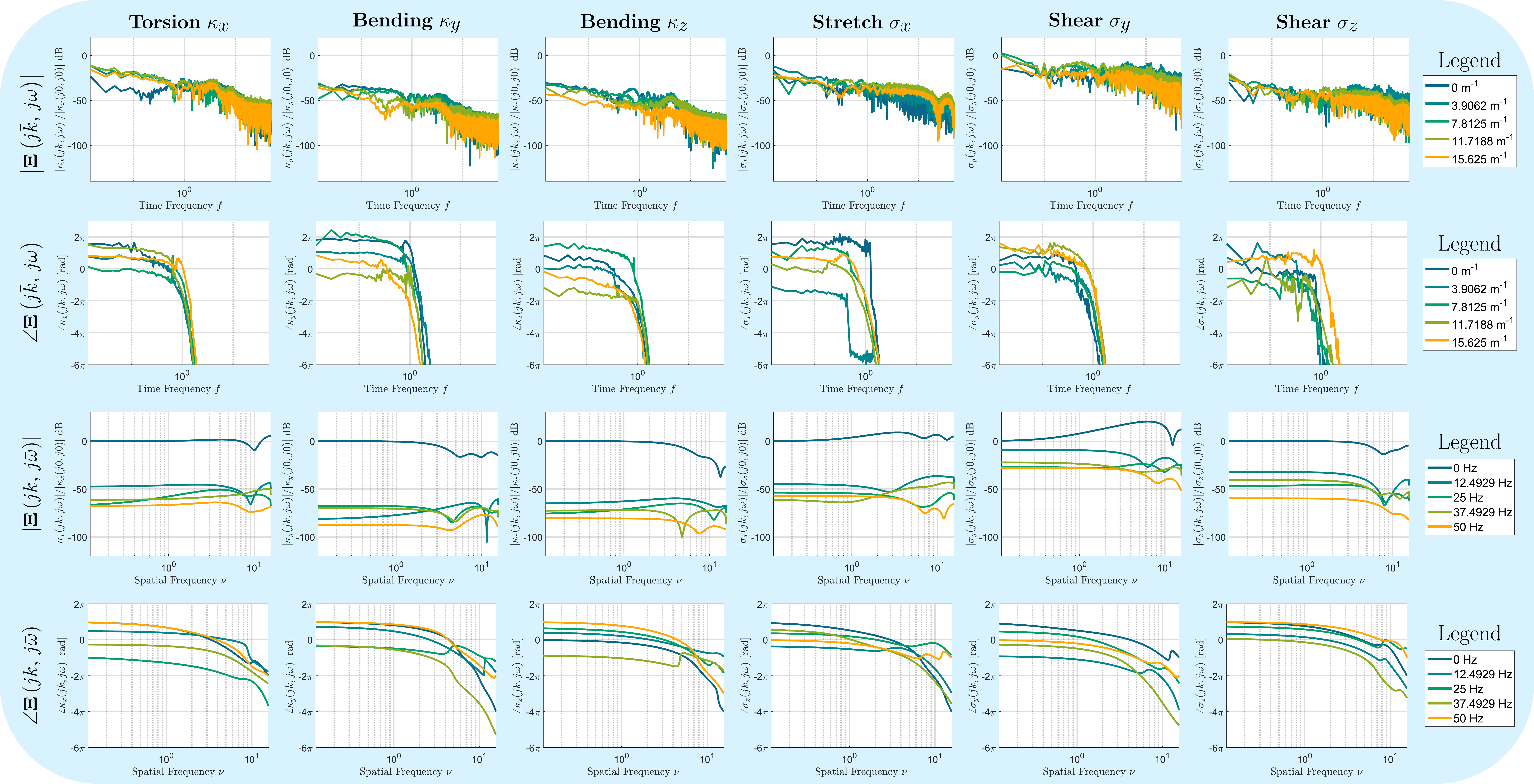}
    \caption{The \ac{STFT} of experimental data from the H-Support prototype. The magnitude values are normalized to $|\Xi_i(j 0, \, j 0)|$. Moreover, zero-padding is applied \textcolor{black}{to the spatial samples, increasing} the resolution of the spatial frequencies.}
    \label{fig:stft_exp}
\end{figure*}

From the samples $\bm{\xi}\left(n \lambda_s, \, m T_s\right)$, it is possible to compute the \ac{STFT} through the \ac{FFT} algorithm. Fig. \ref{fig:stft_exp} shows the spectra of the experimental data, where zero-padding was applied \textcolor{black}{to the spatial samples} to increase the resolution of the wavenumbers.

\subsubsection{Time spectrum of the Spatial Harmonics}
\color{black}
Despite sharing the same body geometry and actuator routing, the experimental spectrum of the H-Support prototype differs from the simulation results.

The torsion spectrum shows a dominance of higher harmonics with respect to the constant component at $\nu = \SI{0}{\meter^{-1}}$. This indicates that torsion is not uniformly distributed along the rod, in contrast to the strain analysis presented in \eqref{eq:implicit_strain}. This can be attributed to the friction of the tendons, which asymmetrically distributed the torsion along its length.

Another major difference between the simulation and the experiment is the presence of antagonistic actuation. In the experimental setup, the pneumatic chambers push the robot while the tendons pull it, resulting in opposing actions. This antagonistic configuration effectively increases the system’s stiffness, thereby limiting the range of deformations typically observed in a cylindrical \ac{CSR}. 
This effect is also reflected in the spectrum. For instance, the constant harmonic in the bending case becomes more pronounced compared to the simulation. This can be explained by the increased global stiffness due to the antagonistic action, which penalizes large and non-uniform bending deformations.

Conversely, the shear spectrum exhibits the opposite trend, where the constant component is less significant compared to the higher harmonics. This behavior can be interpreted as the combined effect of tendon friction and the interaction between the antagonistic actuators.
Finally, discrepancies between the simulated and experimental spectra may also arise from imperfections in the tendon routing.

Finally, the system excited by \eqref{eq:chirp_simulations} exhibits distinct vibration patterns, characterized by a main resonance peak in $\kappa_z$ and $\kappa_x$ modes. 
As highlighted in Sec. \ref{sec:numerical_validation}, the output \ac{STFT} reveals a sequence of resonance and anti-resonance peaks, which correspond to the vibrational modes resulting from the coupling between the various deformation components and the input excitation.
\color{black}

\begin{figure*}
    \centering
    \includegraphics[width=1.0\linewidth]{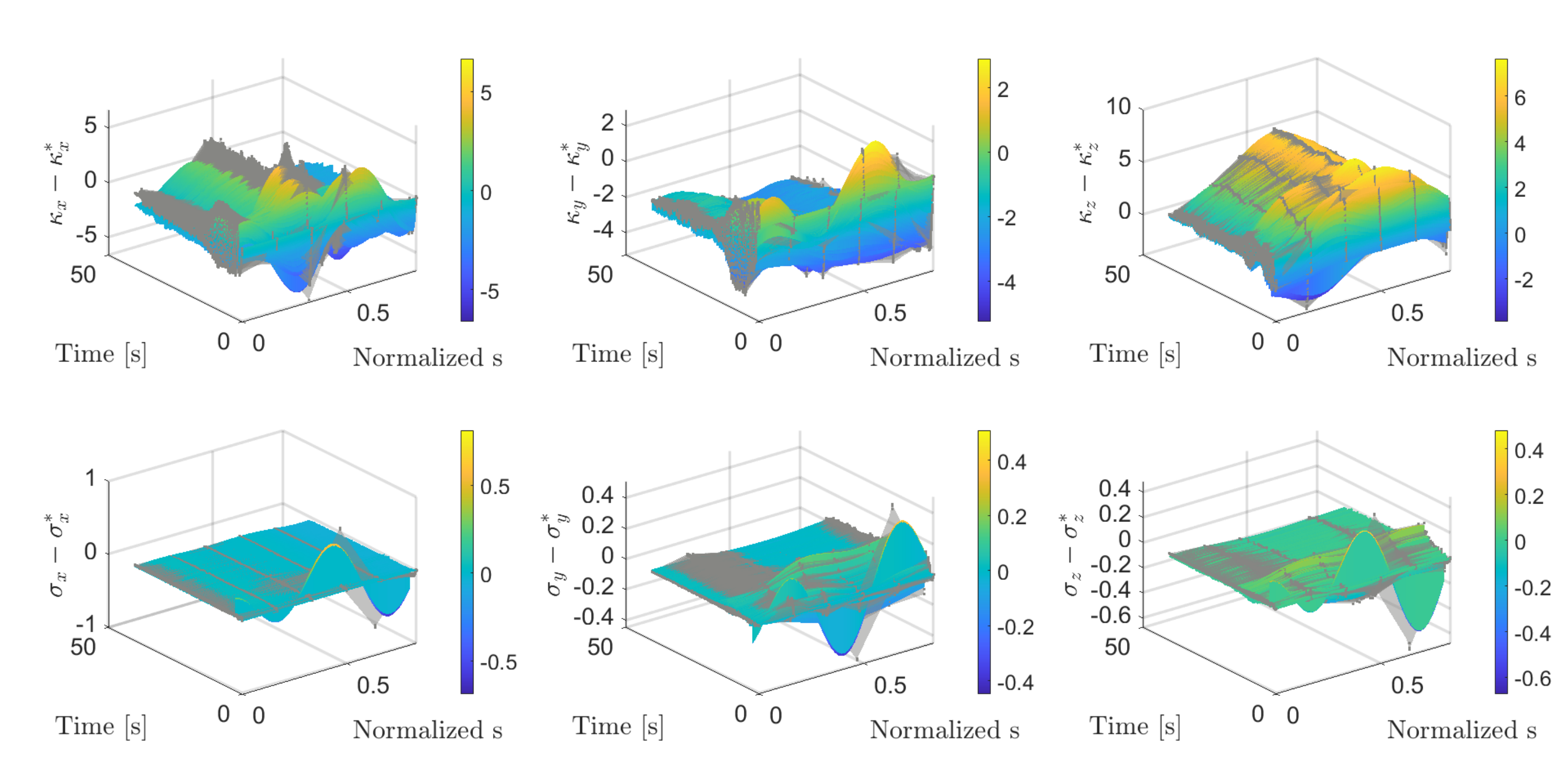}
    \caption{\textcolor{black}{Comparison between the reconstructed strain through \ac{BPD} and the experimental strain. In grey the experimental strain samples.}}
    \label{fig:bpd_st}
\end{figure*}

\subsubsection{Spatial spectrum of the Time Harmonics}
The curves \textcolor{black}{$\bm{\Xi}(jk, \, j \bar{\omega})$} provide insights into the spatial spectrum when the \ac{CSR} is subjected to a specific time-frequency input. 

\color{black}
Across the strain modes, the constant time-harmonic (i.e., \(f = \SI{0}{\hertz}\)) is prominent w.r.t. the other ones. This behavior results from the natural filtering effect typical of mechanical systems, which attenuates high-frequency vibrations.

From these curves, we can derive analogous insights from the perspective of spatial harmonics. In particular, the bending mode shows that the constant component remains as relevant as the harmonics up to approximately $\SI{2}{\meter^{-1}}$. Conversely, the torsion and shear modes exhibit a non-uniform distribution along the rod, where higher spatial harmonics become more pronounced than the constant component.
\color{black}

\subsection{Strain Fitting}
 From the experimental samples, the strain field can be reconstructed using the \ac{BPD} algorithm. The signal dictionary $\bm{B}_{\bm{q}}$ is composed of the polynomial and trigonometric bases.
 Fig. \ref{fig:bpd_st} shows the result of the fitting, both in space and in time. The \ac{BPD} reduces the noise from the strain samples, allowing an accurate representation, and leveraging the signal's sparsity. 
 The displayed results are obtained for the sparsity vector $\bm{\gamma} = [0.5, \,  0.5, \,  0.5, \, 0.07, \,  0.05, \,  0.05]^{\top}$.

\begin{figure*}
    \centering
    \includegraphics[width=1.0\linewidth]{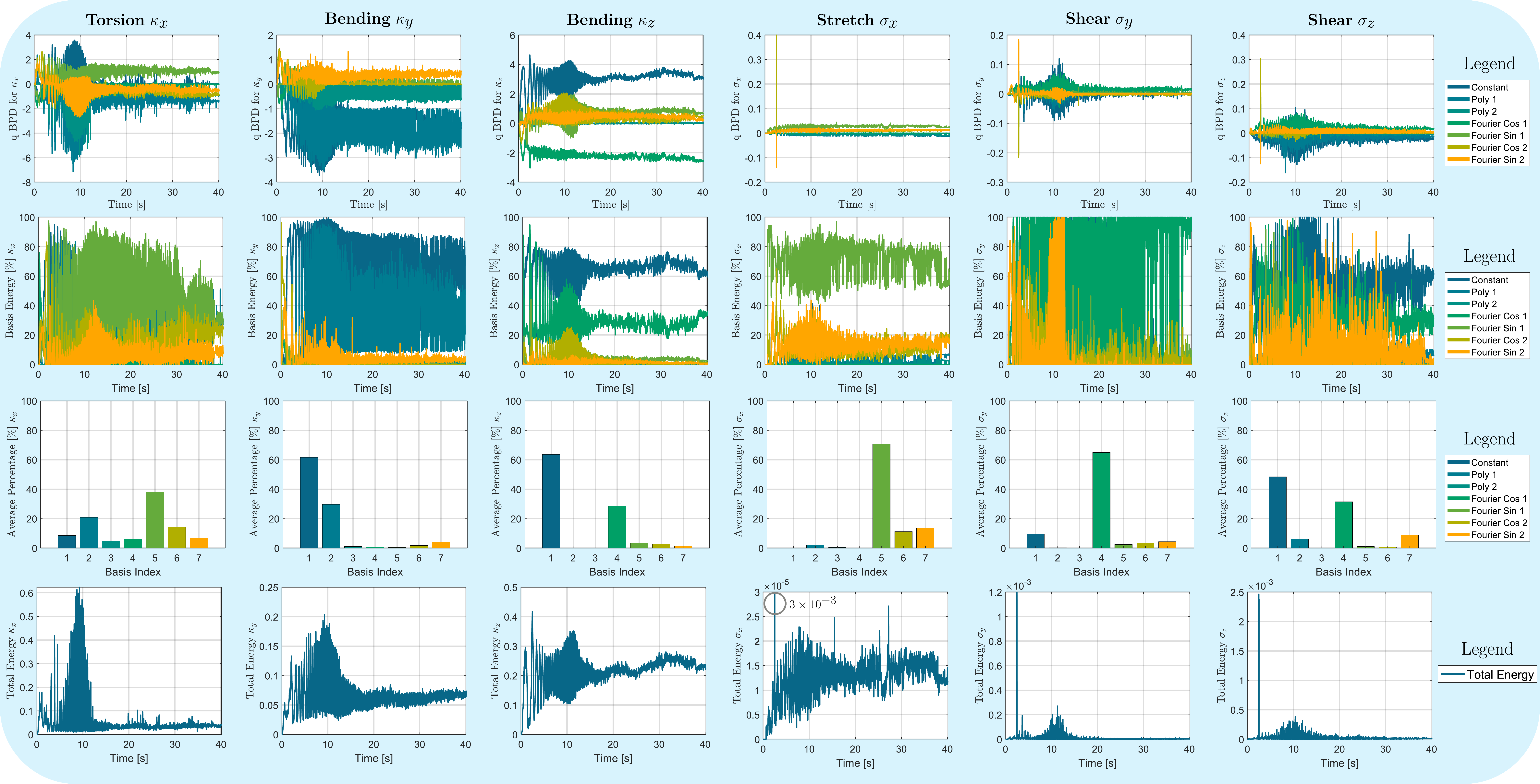}
    \caption{Application of \ac{BPD} to the experimental data. The coefficients $\bm{q}(t)$ are displayed over time in the first row. The energy contribution of each basis over time is shown in the second row, and the average energy contribution value is shown in the third. The total energy of the reconstructed strain is shown in the final row.}
    \label{fig:bpd_q}
\end{figure*}

The first row of Fig. \ref{fig:bpd_q} shows the evolution of the coefficients $\bm{q}$ over time, providing insight into the relevance of each basis. As detailed in Sec. \ref{spectrum_extraction:bpd}, the energy ratio for each basis is calculated and presented in the second row of Fig. \ref{fig:bpd_q}. The third row depicts the average truncation index over time, emphasizing the most significant bases. Lastly, the fourth row illustrates the total energy of the reconstructed signal over time.

The torsion exhibits a diverse combination of basis functions, including both polynomial and trigonometric components, consistent with the spectrum analysis.
For $\kappa_y$, the most dominant basis \textcolor{black}{is} polynomial, specifically the constant and first-order terms. The second-order trigonometric basis ($\nu = \SI{2}{\meter^{-1}}$) plays a notable role, particularly in the first 10 seconds.
Regarding $\kappa_z$, the constant term and the first-order trigonometric basis are the most significant contributors.

The trigonometric bases are mostly used to reconstruct the linear strain modes. 
For the stretch component $\sigma_x$, the constant term is zero, with the first-order trigonometric basis dominating the reconstruction. The shear components $\sigma_y$ and $\sigma_z$ are accurately approximated by first- and second-order trigonometric terms. Unlike the other linear modes, $\sigma_z$ exhibits a significant constant component.
These results align with the previously discussed spectra, where the curves \textcolor{black}{$\bm{\Xi}(jk, \, j \bar{\omega})$} exhibit an increase in magnitude corresponding to these wavenumbers.

\subsection{Comparison between the measured and reconstructed Backbone}
\begin{figure*}
    \centering
    \includegraphics[width=1.0\linewidth]{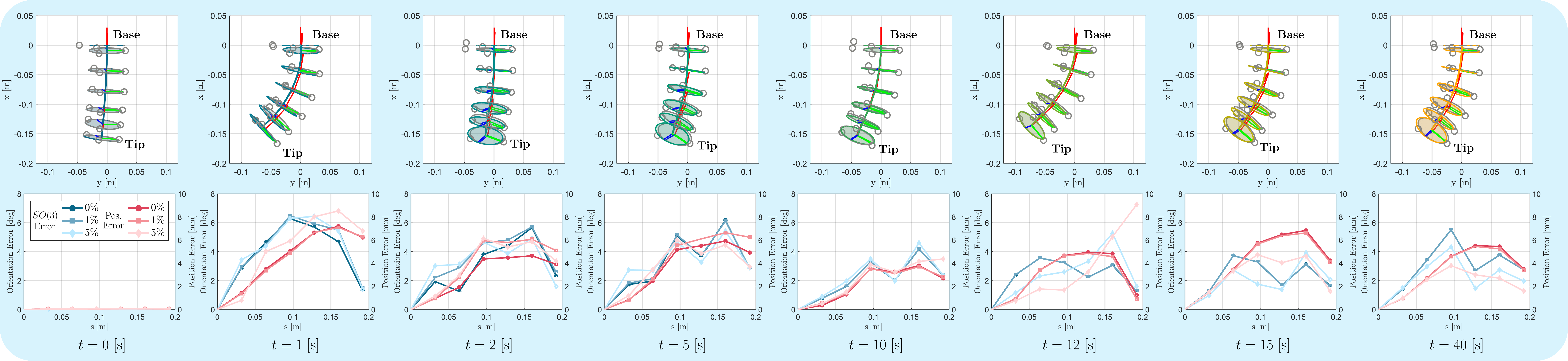}
    \caption{Comparison between the experimental and reconstructed backbone using \ac{BPD}. The last row presents the position and orientation errors, which exhibit an increasing trend along the length of the rod. The errors are computed with the truncation index of $0\%$, $1\%$, and $5\%$.}
    \label{fig:bpd_se3}
\end{figure*}

To evaluate the accuracy of the proposed method, we compare the reconstructed backbone with experimental data obtained from the VICON system. Figure \ref{fig:bpd_se3} shows the reconstructed backbone without truncation (colored) alongside the corresponding experimental pose (gray) for the sensorized cross-sections at eight distinct time instants.

\color{black}
Below each time instant, the orientation and position errors are reported for three truncation thresholds—no truncation ($0\%$), $1\%$, and $5\%$. 
Each basis was assigned an index according to the energy truncation ratio, and all bases with a truncation index below the specified threshold \eqref{eq:truncation_index} are discarded.
\color{black}
The position error is defined as the Euclidean distance between the centers of the reconstructed and experimental cross-sections, while the orientation error is computed using the $\textnormal{dist}_{SO(3)}$ operator \eqref{eq:distso3_definition}. Without truncation, the maximum position error reaches $\SI{7.190}{m\meter}$ (i.e., $3.7448 \% L$), while the maximum orientation error is $6.284^{\circ}$.

Regardless of the truncation threshold, the errors exhibit an increasing trend, which can be attributed to the propagation of fitting errors in the forward kinematics. As these errors accumulate through the integration of the reconstructed strain field, the tip pose is the most affected, reflecting the cumulative effect of all preceding inaccuracies.

In terms of truncation, applying a $1\%$ threshold eliminates $11$ \ac{DoFs}, resulting in a maximum position error of $\SI{7.0872}{m\meter}$ and an orientation error of $6.484^{\circ}$. Increasing the truncation to $5\%$ reduces the model by $23$ \ac{DoFs}, but leading to a higher position error of $\SI{9.076}{m\meter}$ (i.e., $4.7271 \% L$) and a orientation error of $6.443^{\circ}$.

\color{black}
\subsection{External Impulsive Forces}
To demonstrate the effectiveness of the proposed method in handling contact interactions, we conducted two experiments in which our prototype underwent multiple impacts and contacts.

\subsubsection{External Impulsive Force on Passive H-Support}
\begin{figure*}
    \centering
    \includegraphics[width=1.0\linewidth]{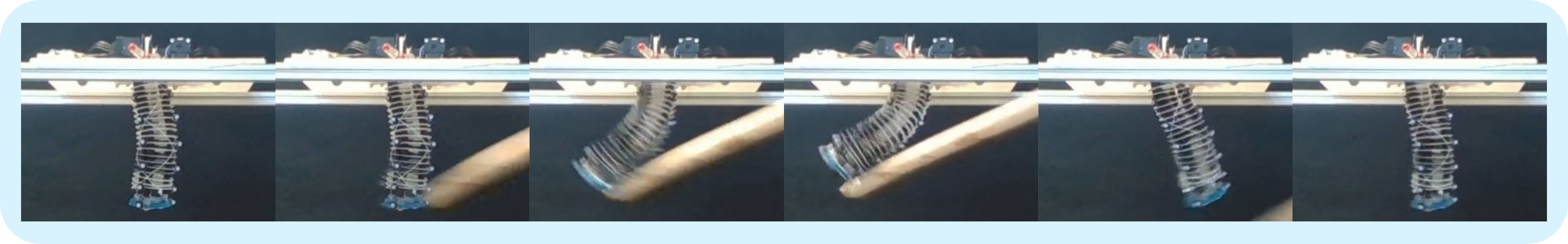}
    \caption{\textcolor{black}{Frames of the external impulsive force on the passive H-Support.}}
    \label{fig:contact_frames}
\end{figure*}

\begin{figure}
    \centering
    \includegraphics[width=1.0\linewidth]{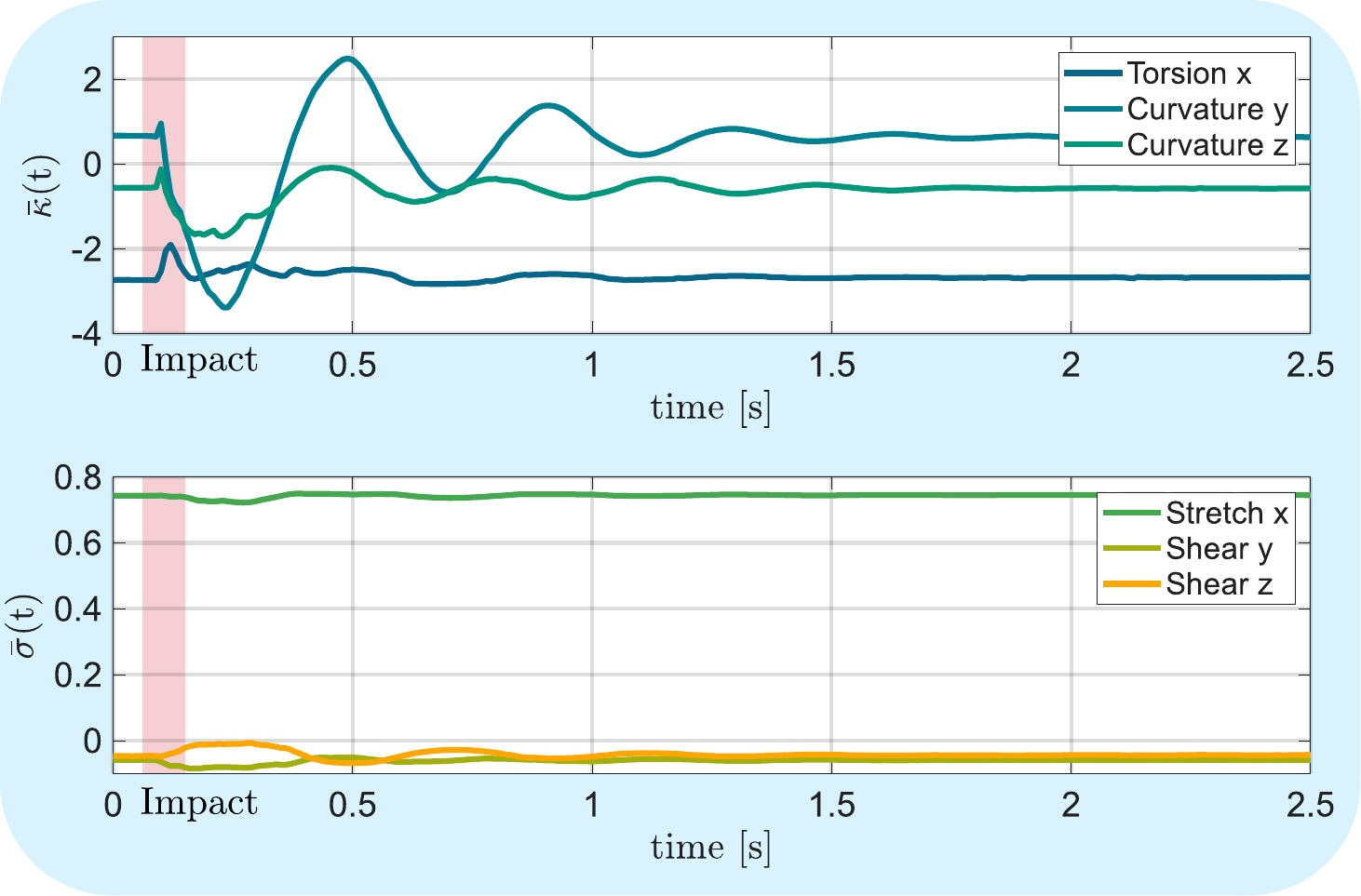}
    \caption{\textcolor{black}{Time evolution of the average strain during the external impulsive force experiment.}}
    \label{fig:avg_xi_woodstick}
\end{figure}

In the first experiment, the unactuated robot was struck by an external object to observe its autonomous response to an external impulse. Snapshots of the experiment are shown in Fig. \ref{fig:contact_frames}, while Fig. \ref{fig:avg_xi_woodstick} presents the spatially averaged strain, illustrating the system’s reaction to the applied force.

\begin{figure*}
    \centering
    \includegraphics[width=1.0\linewidth]{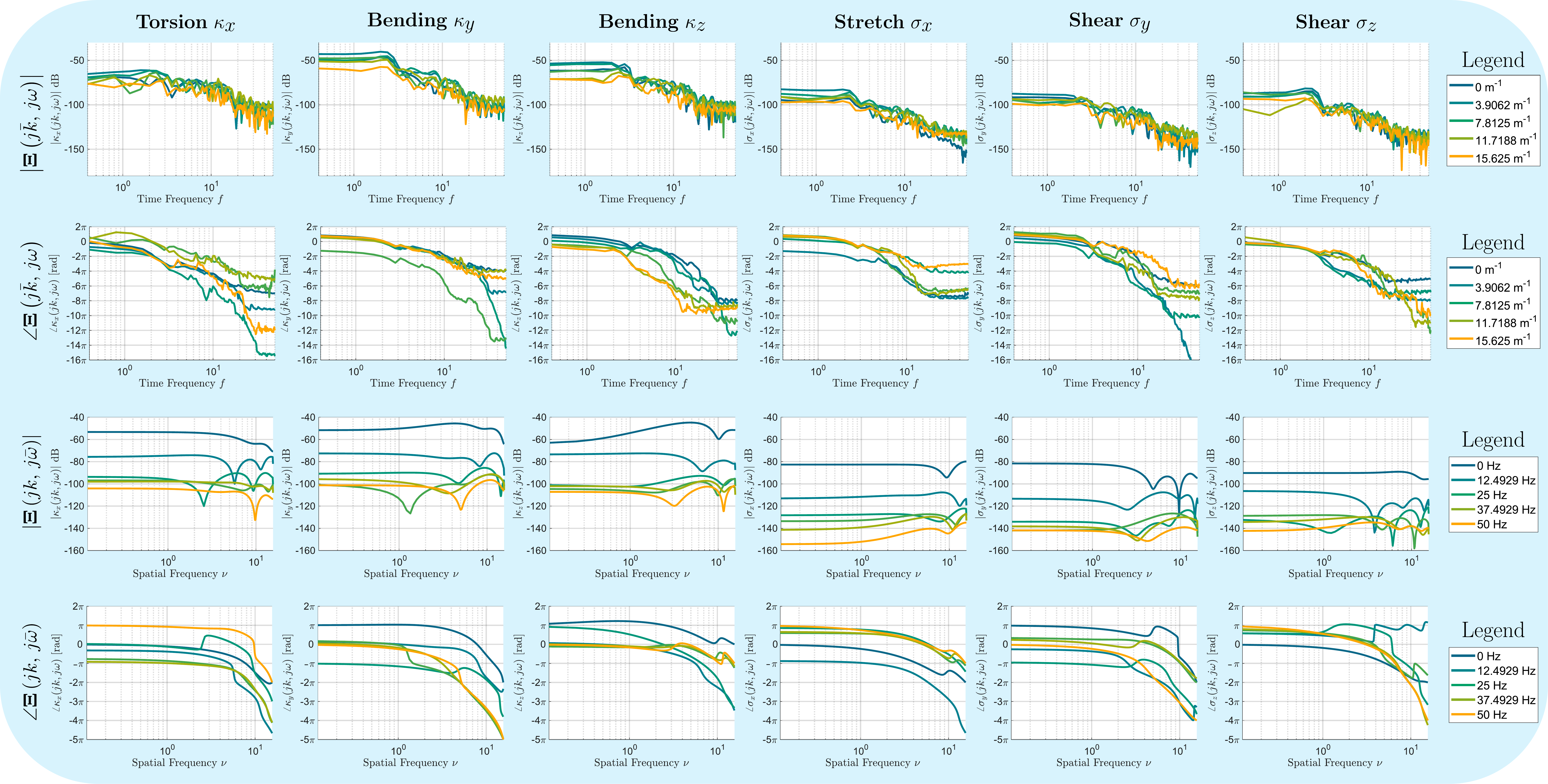}
    \caption{\textcolor{black}{\ac{STFT} of the passive H-Support perturbed by external impulsive force.}}
    \label{fig:woodstick_spectrum}
\end{figure*}

From the strain samples $\bm{\xi}\left(n \lambda_s, m T_s \right)$, we can compute the \ac{STFT}, represented in Fig. \ref{fig:woodstick_spectrum}.

The strain mode that primarily responds to the external force is the bending mode, which exhibits a second-order mechanical frequency response, most prominently in $\kappa_y$ and less evident in $\kappa_z$. 
The shear strain also reacts to the impulse with a similar dynamic behavior. 

The external force not only induces an impulsive dynamic response but also excites new spatial harmonics due to the localized nature of the impact.
Therefore, from curves $\bm{\Xi}\left(j k, \, j \bar{\omega}\right)$, new spatial harmonics appear in the range of $6-8 \; \textnormal{m}^{-1}$, revealing the emergence of additional deformation modes following the impact.
Furthermore, this indicates the presence of non-uniform deformation induced by the external force, which is also evident in the experimental snapshots shown in Fig. \ref{fig:contact_frames}.

\subsubsection{External Impulsive Force on Actuated H-Support}
In the second experiment, we actuated our prototype using the same input law described in Sec. \ref{sec:experimental_validation}. 
During the system’s evolution, we applied impulsive forces to the active robot, inducing deformations different from those normally generated by the actuation. 
We then compared the resulting spectra, examining cases in which the H-Support prototype was subjected to these impulsive forces versus when it was not.

\begin{figure}
    \centering
    \includegraphics[width=1.0\linewidth]{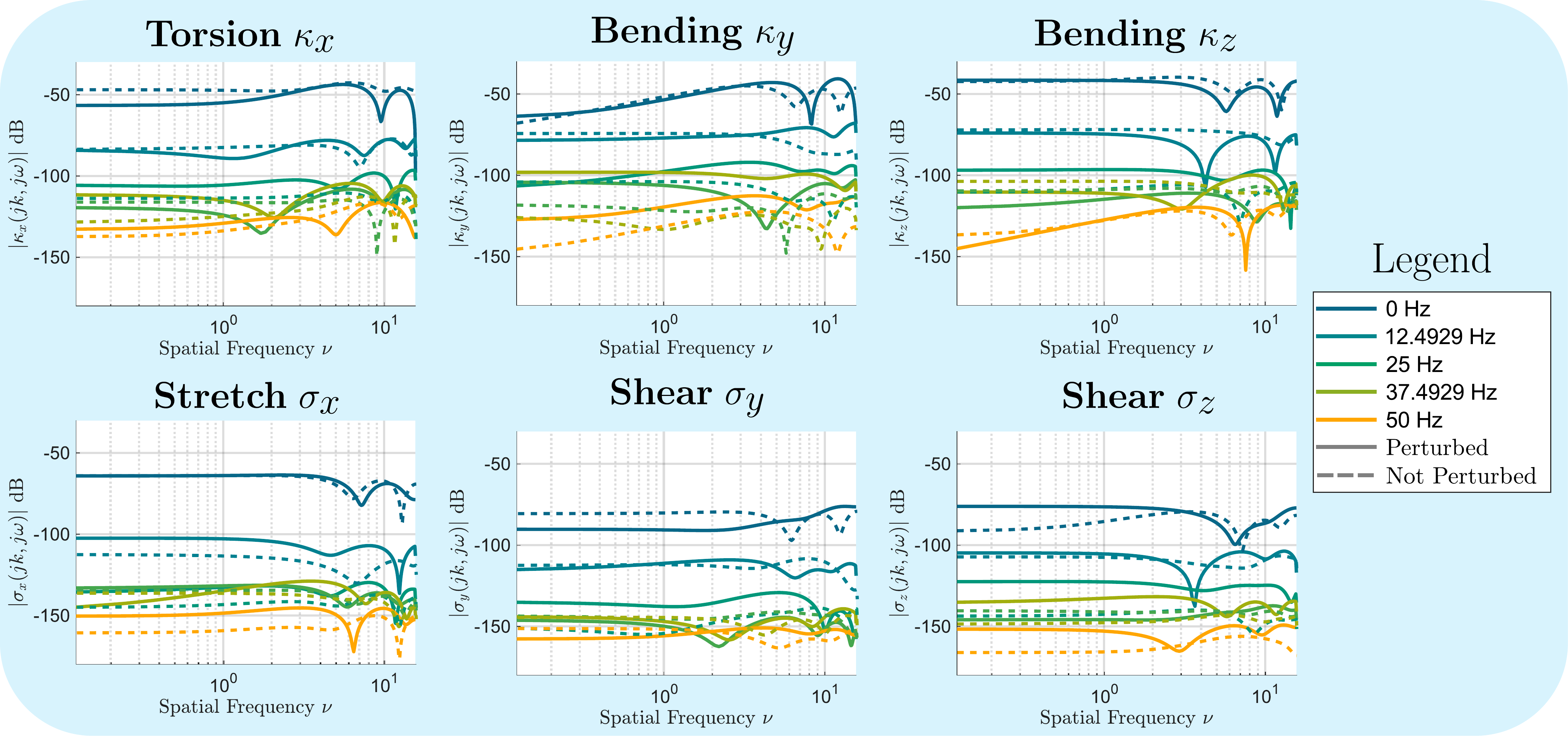}
    \caption{\textcolor{black}{\ac{STFT} of the H-Support subjected to a chirp input, perturbed by an external impulsive force (solid line). The unperturbed experimental spectrum is shown for comparison (dotted line).}}
    \label{fig:chirp_contact}
\end{figure}

Of particular interest are the curves $\bm{\Xi}\left(j k \, , j \bar{\omega}\right)$, which highlight differences in the excited spatial harmonics, as shown in Fig. \ref{fig:chirp_contact}. Similar to the passive case, the perturbations primarily affect the bending and shear modes. Notably, the shear modes exhibit altered spatial harmonics, while the bending modes show slightly different dominant harmonics compared to the unperturbed case.

The spectral differences also vary in the time-frequency domain, indicating that the impulsive perturbations interact with the system’s dynamic response under chirp excitation. Even the torsion and stretch modes show spectral variations between the perturbed and unperturbed cases, suggesting the influence of these perturbations on the axial deformation modes.

From this spectral analysis, the user can construct a signal dictionary to determine the optimal basis for modeling all three components described in \eqref{eq:implicit_strain}, namely static, dynamic, and external contributions.
\color{black}
\section{Conclusion} \label{sec:conclusions}
    This work presented a novel perspective on modeling \acp{CSR} by leveraging the Fourier transform within the framework of \ac{CRT}. By treating the backbone of the robot as a spatial signal, we provided a compact and efficient representation of its deformation, offering a unifying approach to existing modeling strategies. The application of the Fourier transform enhances the understanding of traditional heuristic methods. Furthermore, it facilitates the development of a data-driven methodology for selecting the optimal model in terms of accuracy and computational efficiency.

The proposed framework was validated through both numerical simulations and real-world experiments, confirming its effectiveness in reducing the degrees of freedom while preserving accuracy.

For future work, the spatial spectrum can be applied to both model-based and learning-based controllers as a dense representation of the robot's deformation. 
\color{black}
Another direction is the real-time adaptation of the framework, which would involve updating the optimal basis over time to account for changing conditions, such as external contacts or unmodeled dynamics.
\color{black}
Additionally, a deeper analysis of the system's frequency response could lead to the development of an identification algorithm.

\section*{Appendix} \label{sec:appendix}
\setcounter{equation}{0} 
\renewcommand{\theequation}{A.\arabic{equation}}

\subsection*{Lie Algebra}
\begin{equation} \label{eq:Adjoint_definition}
    \textnormal{Ad}_{\bm{g}} = \begin{bmatrix}
                                    \bm{R} & \bm{0}_{3 \times 3} \\
                                    \tilde{\bm{r}} \bm{R} & \bm{R}
                                \end{bmatrix} \, ,
\end{equation}
\begin{equation} \label{eq:coAdjoint_definition}
\textnormal{Ad}^{*}_{\bm{g}} = \begin{bmatrix}
                                    \bm{R} & \tilde{\bm{r}} \bm{R} \\
                                    \bm{0}_{3 \times 3} & \bm{R}
                                \end{bmatrix} \, ,
\end{equation}
\begin{equation} \label{eq:adjoint_definition}
    \textnormal{ad}_{\bm{\xi}, \, \bm{\eta}} = \begin{bmatrix}
      \tilde{\bm{\kappa}}, \, \tilde{\bm{w}} & \bm{0}_{3 \times 3} \\
      \tilde{\bm{\sigma}}, \, \tilde{\bm{v}} & \tilde{\bm{\kappa}}, \, \tilde{\bm{w}}
    \end{bmatrix} \, ,
\end{equation}
\begin{equation} \label{eq:coadjoint_definition}
    \textnormal{ad}^{*}_{\bm{\xi}, \, \bm{\eta}} = \begin{bmatrix}
      \tilde{\bm{\kappa}}, \, \tilde{\bm{w}} & \tilde{\bm{\sigma}}, \, \tilde{\bm{v}}  \\
      \bm{0}_{3 \times 3} & \tilde{\bm{\kappa}}, \, \tilde{\bm{w}}
    \end{bmatrix} \, ,
\end{equation}
\begin{equation} \label{eq:logSE3_definition}
    \begin{split}
        &\log_{SE(3)} \left(\bm {g}\right) = \beta \left( \alpha_0 \bm{I}_4 - \alpha_1 \bm{g} + \alpha_2 \bm{g}^2 - \alpha_3 \bm{g}^3\right) \, ,
    \end{split}
\end{equation}
where $\beta = \frac{1}{8}\csc ^{3}\left(\frac{\theta}{2}\right)\sec \left(\frac{\theta}{2}\right)$, $\alpha_0 = \theta \cos \left(2\theta \right) - \sin \left(\theta \right)$, $\alpha_1 = \theta \cos \left(\theta \right)+2\theta \cos \left(2\theta \right)-\sin \left(\theta \right)-\sin \left(2\theta \right)$, $\alpha_2 = 2\theta \cos \left(\theta \right)+\theta \cos \left(2\theta \right)-\sin \left(\theta \right)-\sin \left(2\theta \right)$, $\alpha_3 = \theta \cos \left(\theta \right)-\sin \left(\theta \right)$, and
$\theta = \arccos\left(\frac{1}{2} \mathrm{tr}\left(\bm{g}\right) - 1\right)$.
\begin{equation} \label{eq:distso3_definition}
    \textnormal{dist}_{SO(3)}\left(\bm{R}_1, \bm{R}_2\right) = \abs{\arccos\left[\frac{1}{2}\left(\mathrm{tr}\left(\Delta\bm{R}\right) - 1\right)\right]} ,
\end{equation}
where $\Delta\bm{R} = \bm{R}_1^{\top} \bm{R}_2$.
\begin{equation} \label{eq:distse3_defintion}
    \textnormal{dist}_{SE(3)}\left(\bm{g}_1, \, \bm{g}_2\right) = \norm{\log_{SE(3)}\left(\bm{g}_{1}^{-1} \bm{g}_2\right)^{\vee}}_2 \, ,
\end{equation}

\color{black}
\subsection*{Linearization of \ac{GVS} Dynamics}

Let $\bm{q} \in \mathbb{R}^{n_q}$ denote the configuration of a \ac{CSR} modeled using the \ac{GVS} approach, such that $\bm{\xi}(s, t) = \bm{B}_{\bm{q}}(s) \, \bm{q} + \bm{\xi}^{*}(s)$.

The Forward Dynamics can always be expressed in Lagrangian form as
\begin{equation*}
    \ddot{\bm{q}} = - \bm{M}^{-1} \Big( \bm{C}\dot{\bm{q}} + \bm{G} + \bm{K} \bm{q} + \bm{D} \dot{\bm{q}} - \bm{B} \bm{\tau} \Big) 
    = \text{FD}(\bm{q}, \dot{\bm{q}}, \bm{\tau}) \,,
\end{equation*}
where $\bm{M} \in \mathbb{R}^{n_q \times n_q}$ is the mass matrix, $\bm{C} \in \mathbb{R}^{n_q \times n_q}$ the Coriolis matrix, $\bm{G} \in \mathbb{R}^{n_q}$ the gravity vector, $\bm{K}, \bm{D} \in \mathbb{R}^{n_q \times n_q}$ the stiffness and damping matrices, and $\bm{B} \in \mathbb{R}^{n_q \times m}$ the actuation matrix.

Defining the state and input vectors as $\bm{x} = \begin{bmatrix} \bm{q}^{\top} & \dot{\bm{q}}^{\top} \end{bmatrix}^{\top}$, and $\bm{u} = \bm{\tau}$,
the system can be written in state-space form as
\begin{equation*}
    \dot{\bm{x}} = \bm{f}(\bm{x}, \bm{u}) \,, \quad 
    \bm{f}(\bm{x}, \bm{u}) = \begin{bmatrix} 
        \bm{x}_2 \\ \text{FD}(\bm{x}, \bm{u}) 
    \end{bmatrix} \,.
\end{equation*}
Linearization around an equilibrium point $\bar{\bm{x}}_e, \bar{\bm{u}}_e$ yields
\begin{equation} \label{eq:linearized_system}
    \dot{\tilde{\bm{x}}} = \bm{A}_{\text{lin}} \tilde{\bm{x}} + \bm{B}_{\text{lin}} \tilde{\bm{u}} \,,
\end{equation}
where $\tilde{\bm{x}} = \bm{x} - \bar{\bm{x}}_e$ and $\tilde{\bm{u}} = \bm{u} - \bar{\bm{u}}_e$, with
\begin{equation} \label{eq:A_lin}
    \bm{A}_{\text{lin}} = 
    \left. \begin{bmatrix}
        \bm{0} & \bm{I} \\
        - \bm{M}^{-1} \left( \bm{K} + \frac{\partial \bm{G}}{\partial \bm{q}} \right) & - \bm{M}^{-1} \bm{D}
    \end{bmatrix} \right|_{\bm{x} = \bar{\bm{x}}_e} \,,
\end{equation}
\begin{equation} \label{eq:B_lin}
    \bm{B}_{\text{lin}} = 
    \left. \begin{bmatrix}
        \bm{0} \\ 
        \bm{M}^{-1} \bm{B} 
    \end{bmatrix} \right|_{\bm{x} = \bar{\bm{x}}_e} \,.
\end{equation}

The equilibrium $(\bar{\bm{x}}_e, \bar{\bm{u}}_e)$ is obtained by solving
\begin{equation*}
    \bm{f}(\bar{\bm{x}}_e, \bar{\bm{u}}_e) = \bm{0} \,,
\end{equation*}
which corresponds to the static solution of the system.
\color{black}

\subsection*{Functional Basis for \ac{GVS}}
\textit{Polynomial Basis}:
\begin{equation} \label{eq:polynomial_basis}
    \xi_i(s, t) - \xi_i^{*}(s) = \sum_{h = 0}^{n - 1} q_h s^{h} \, ,
\end{equation}
\textit{Fourier Basis}:
\begin{equation} \label{eq:fourier_basis}
    \xi_i(s, t) - \xi_i^{*}(s) = \sum_{h = 0}^{n - 1} q_{h, 1} \cos(h k) + q_{h, 2} \sin(h k) \, ,
\end{equation}
\textit{Gaussian Basis}:
\begin{equation} \label{eq:gaussian_basis}
    \xi_i(s, t) - \xi_i^{*}(s) = \sum_{h = 0}^{n - 1} q_{h} \exp\left(- \frac{\left(s - h/n\right)^{2}}{c^2} \right) \, ,
\end{equation}
where $c = \left(2 \sqrt{\ln\left(2\right)} n\right)^{-1}$.

\subsection*{Properties of Fourier Transform}
\begin{equation} \label{eq:fourier_definition}
    \fourier{f(t)} = F(j \omega) = \int_{-\infty}^{+\infty} f(t) \, e^{- j \omega t} \, \textnormal{d} t \, ,
\end{equation}
\color{black}
\begin{equation} \label{eq:fourier_linearity}
    \fourier{\alpha f(t) + \beta g(t)} = \alpha F(j \omega) + \beta G(j \omega) \, ,
\end{equation}
\begin{equation} \label{eq:fourier_delay}
    \fourier{f(t - T_0)} = F(j \omega) e^{-j \omega T_0} \, ,
\end{equation}
\begin{equation} \label{eq:fourier_diff}
    \fourier{\dot{f}(t)} = j \omega \, F(j \omega) \, ,
\end{equation}
\begin{equation} \label{eq:fourier_conv}
    \fourier{f(t) \cdot g(t)} = F(j \omega) \ast G(j \omega) \, ,
\end{equation}
\begin{equation} \label{eq:fourier2d_product}
    \fourier{f(s, t) = g(s) \cdot h(t)} = G(j k) \cdot H(j \omega) \, ,
\end{equation}
\begin{equation} \label{eq:fourier_parseval}
    \int_{-\infty}^{+\infty} \abs{f(t)}^2 \, \textnormal{d}t = \frac{1}{2 \pi} \int_{-\infty}^{+\infty} \abs{F(j \omega)}^2 \, \textnormal{d} \omega \, ,
\end{equation}
\color{black}
\begin{equation} \label{eq:discrete_parseval}
    \sum_{n = 0}^{N - 1} \abs{f(n)}^2 = \frac{1}{N} \sum_{n = 0}^{N - 1} \abs{F(n)}^2 \, .
\end{equation}
\color{black}

\color{black}
\subsection*{Notable Functions}
\begin{equation} \label{eq:length_space_window}
    \Pi_L(s) = \rect{\frac{s - L/2}{L}} \, ,
\end{equation}
 where $\rect{\cdot}$ is the rectangular function \cite[Ch. 4]{bracewell2007fourier}.
\begin{equation} \label{eq:sft_length_window}
    \Pi_L(jk) = L \, \sinc{k \frac{L}{2}} e^{-j k \pi L} \, ,
\end{equation}
where $\sinc{\cdot}$ is the cardinal sine.
\begin{equation} \label{eq:piece_window}
    \Pi_h\left(s\right) =  \rect{\frac{s - \left(h + \frac{1}{2}\right) \lambda_p}{\lambda_p}} \, ,
\end{equation}
\begin{equation} \label{eq:sft_piece_window}
    \Pi_h\left(jk\right) = \lambda_p \, \sinc{k \frac{\lambda_p}{2}} e^{-j k \left(h + \frac{1}{2}\right)\lambda_p} \, ,
\end{equation}
\begin{equation} \label{eq:zoh_pcs}
    \bm{H}_0(jk) =  \lambda_p \left(\sinc{k \frac{\lambda_p}{2}} e^{-j k \frac{\lambda_p}{2}}\right) \bm{I}_{6} \, ,
\end{equation}
\begin{equation} \label{eq:foh_pas}
    \bm{H}_1(jk) =  \lambda_p \, \sinc{k \frac{\lambda_p}{2}}^2 \left(1 + jk \lambda_p\right) e^{-j k \frac{\lambda_p}{2}} \bm{I}_{6} \, .
\end{equation}
\color{black}

\bibliographystyle{sageh}
\bibliography{biblio}

\end{document}